\documentclass[10pt,journal,compsoc]{IEEEtran}
\usepackage{algorithm} 
\usepackage{algorithmic}
\usepackage[english]{babel}
\usepackage{amsmath}
\usepackage{amssymb}
\usepackage{multirow}
\usepackage{textcomp}
\usepackage{epsfig}
\usepackage{bm}
\usepackage{rotating}
\usepackage{color}
\usepackage[pagebackref=true,breaklinks=true,letterpaper=true,colorlinks,bookmarks=false]{hyperref}
\usepackage{breakurl}
\usepackage{booktabs}
\makeatletter

\ifCLASSOPTIONcompsoc
  \usepackage[nocompress]{cite}
\else
  \usepackage{cite}
\fi
\ifCLASSINFOpdf
\else
\fi

\begin{document}
%
\title{Recognizing Partial Biometric Patterns}
\author{Lingxiao He, \IEEEmembership{Student Member, IEEE}, Zhenan Sun, \IEEEmembership{Member, IEEE}, Yuhao Zhu and Yunbo Wang
\IEEEcompsocitemizethanks{\IEEEcompsocthanksitem L. He, Z. Sun (corresponding author), Y. Wang are with the Center for Research
on Intelligent Perception and Computing (CRIPAC), National Laboratory of Pattern
Recognition (NLPR), Institute of Automation, CAS Center for Excellence in Brain
Science and Intelligence Technology, Chinese Academy of Sciences University (UCAS) of Chinese Academy of Sciences.
E-mail: {lingxiao.he, znsun, yunbo.wang}@nlpr.ia.ac.cn. Y. Zhu is with the Center for Research on Intelligent Perception and Computing (CRIPAC).
E-mail:yuhao.zhu@cripac.ia.ac.cn.

\IEEEcompsocthanksitem This work is funded by the National Natural Science Foundation of China (Grant No. 61427811, 61573360) and the National Key Research and Development Program of China (Grant No. 2017YFC0821602, 2016YFB1001000).
}}

\markboth{Journal of \LaTeX\ Class Files,~Vol.~14, No.~4, August~2018}%
{Shell \MakeLowercase{\textit{et al.}}: Bare Demo of IEEEtran.cls for Computer Society Journals}
\IEEEtitleabstractindextext{%
\begin{abstract}
    Biometric recognition on partial captured targets is challenging, where only several partial observations of objects are available for matching. In this area, deep learning based methods are widely applied to match these partial captured objects caused by occlusions, variations of postures or just partial out of view in person re-identification and partial face recognition. However, most current methods are not able to identify an individual in case that some parts of the object are not obtainable, while the rest are specialized to certain constrained scenarios. To this end, we propose a robust general framework for arbitrary biometric matching scenarios without the limitations of alignment as well as the size of inputs. We introduce a feature post-processing step to handle the feature maps from FCN and a dictionary learning based Spatial Feature Reconstruction (SFR) to match different sized feature maps in this work. Moreover, the batch hard triplet loss function is applied to optimize the model. The applicability and effectiveness of the proposed method are demonstrated by the results from experiments on three person re-identification datasets (Market1501, CUHK03, DukeMTMC-reID), two partial person datasets (Partial REID and Partial iLIDS) and two partial face datasets (CASIA-NIR-Distance and Partial LFW), on which state-of-the-art performance is ensured in comparison with several state-of-the-art approaches. The code is released online and can be found on the website: \url{https://github.com/lingxiao-he/Partial-Person-ReID}.
\end{abstract}

}
\maketitle

\IEEEdisplaynontitleabstractindextext

\IEEEpeerreviewmaketitle

\IEEEraisesectionheading{\section{Introduction}\label{sec:introduction}}

\IEEEPARstart{B}{iometric} recognition, especially face recognition and person re-identification (re-id), has attracted significant research attention as the demand of identification using images captured by CCTV cameras and video surveillance systems growing rapidly.
In these scenarios, the random poses and perspectives of the target object, unwilling occlusions caused by other objects (e.g. hair, sunglasses even other individuals for a person or eyelids/eyelashes for irises) and only partly captured images of target objects would degrade the performance of surveillance systems.
With the great progress has been made on biometric identification in recent years due to the development of deep learning, many approaches are proposed from global researches. And we consider these approaches can be divided into two generations.

\begin{figure}[t]
    \centering
       \vspace{0em}
    \includegraphics[width=7.5cm]{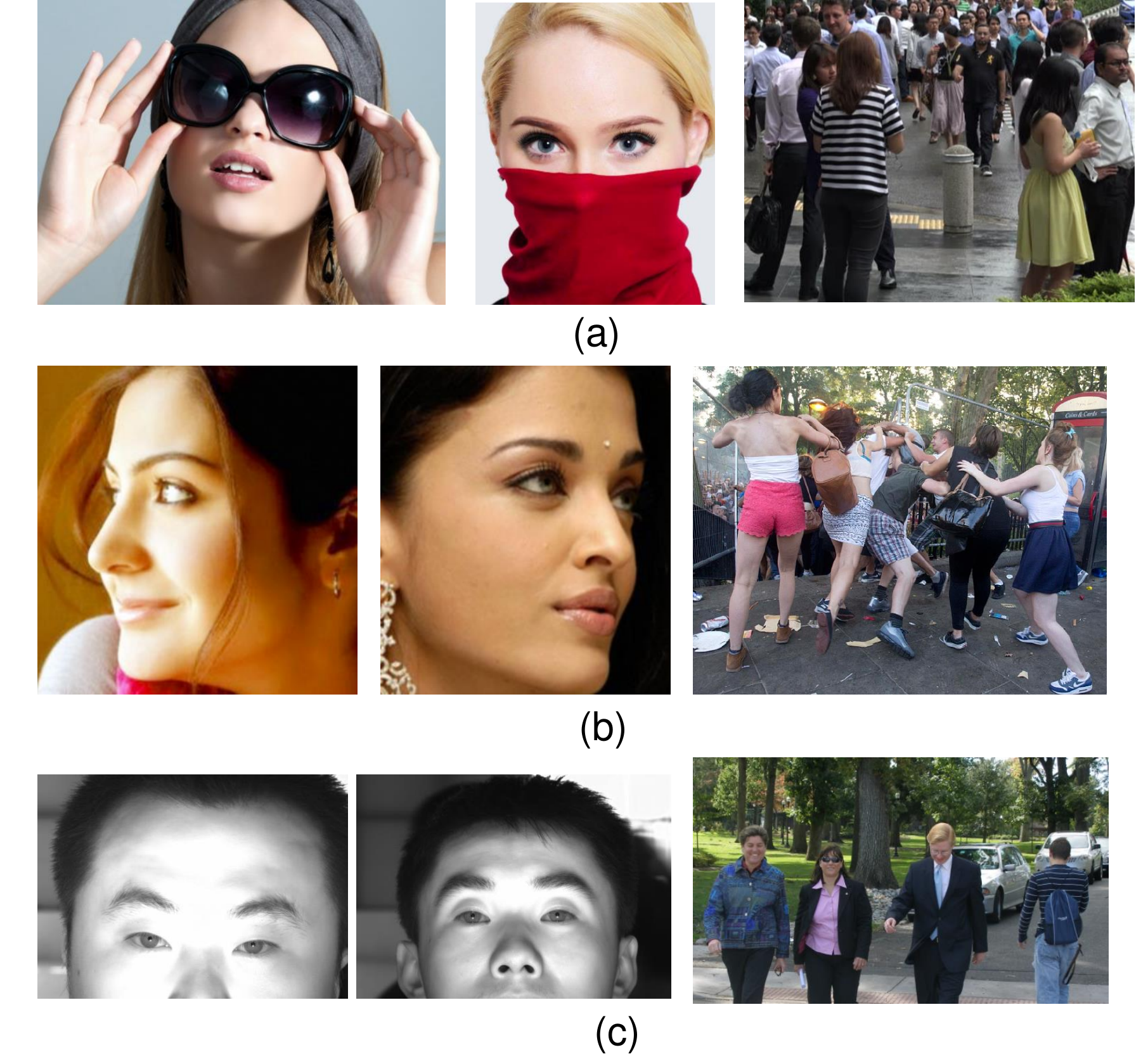}
     \caption{Examples of captured images in the real-world scenarios. (a) a face is
easily occluded by accessories such as sunglasses, scarfs and a person is occluded by other persons; (b) arbitrary posture of object; (c) an object may be positioned partially outside cameras view. }
    \label{fig1}
\end{figure}

The first generation approaches generally assume that each image covers full glance of one object. However, the assumption of biometric matching on full and frontal images does not always hold in real-world scenarios, where we merely have access to a few parts of images for identification. For instance shown in Fig. \ref{fig1}, a face is easily occluded by accessories such as sunglasses, scarfs, and a person on the street can easily be occluded by moving obstacles (e.g., cars, other persons) and static ones (e.g., trees, barriers), resulting in partial observations of the target object. Besides, the frequently presented arbitrary posture of an object in video surveillance introduces additional difficulties to real-world biometric identification problems. Moreover, an object may be positioned partially outside cameras view, resulting in an arbitrary-size image. These emerging problems would reduce the performance of the first generation methods.
\begin{figure}[t]
    \centering
       \vspace{0em}
    \includegraphics[width=8.8cm]{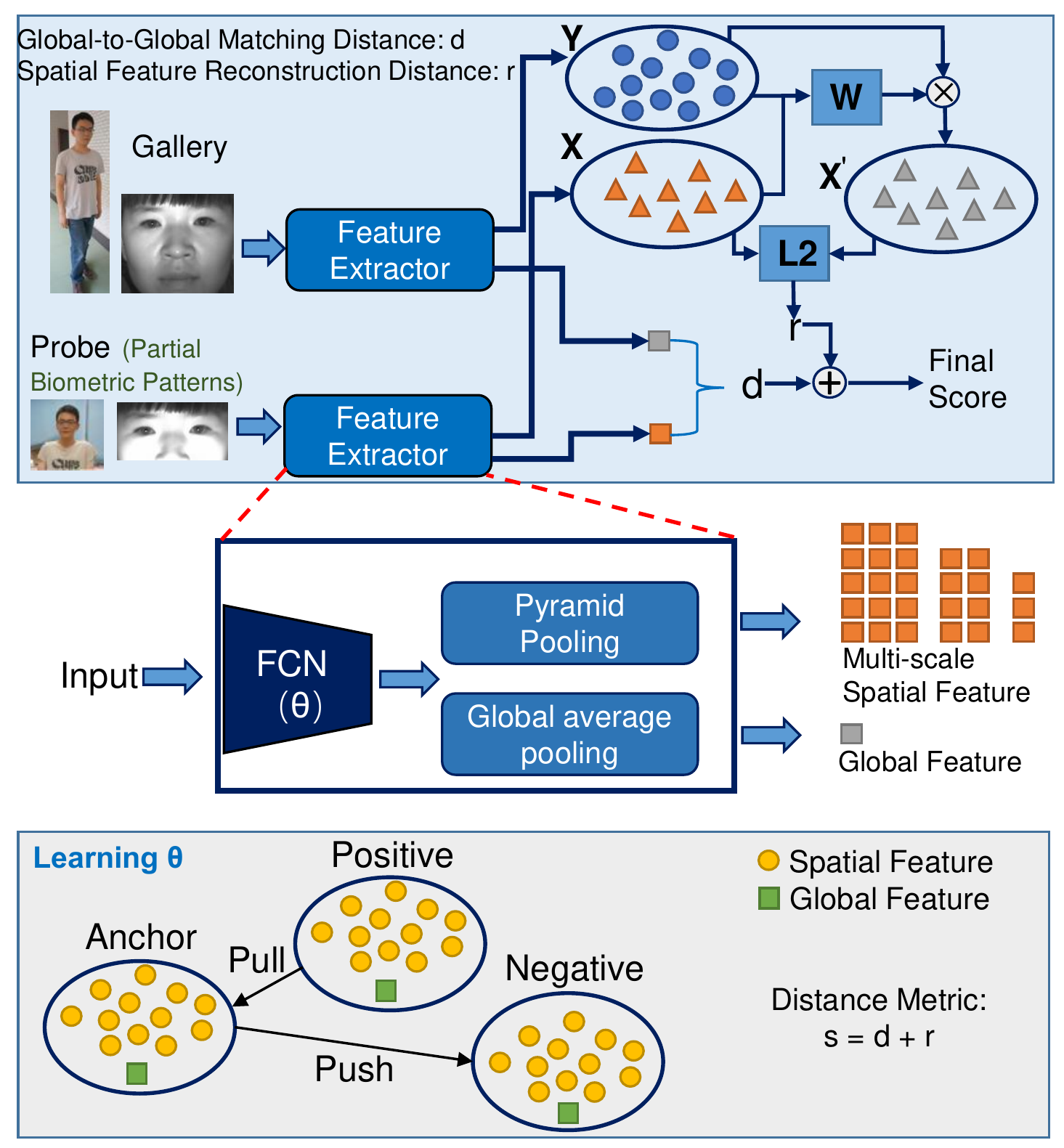}
     \caption{The proposed framework. Fully Convolutional Network (FCN) is utilized to generate spatial feature maps of a certain size. And then a feature post-processing unit consist of global averaging pooling and pyramid pooling is utilized to produce global feature and multi-scale spatial feature. For a probe and a gallery, we first extract their spatial feature $\mathbf{X}$ and $\mathbf{Y}$ and global feature. Secondly, we calculate the reconstruction coefficients $\mathbf{W}$, and then obtain the reconstruction spatial feature $\mathbf{X}^{'}$ by $\mathbf{YW}$. Finally, we fuse the global-to-global matching distance $d$ and spatial feature reconstruction (SFR) distance $r$. As shown in the bottom of figure, we use the triplet loss to optimize $\theta$ by using the distance metric: $s=d+r$. }
    \label{fig2}
\end{figure}

The drawbacks of first generation approaches makes researchers to design a framework to address partial biometric identification problems, where the second generation approaches advent. To match an arbitrary patch of an image, some researchers resort to re-scale an arbitrary patch of the image to a fixed-size image. However, the performance would be significantly degraded due to the undesired deformation. Part-based models \cite{chen2016similarity}, \cite{cheng2016person}, \cite{gutta2002investigation}, \cite{li2017learning}, \cite{pan2007part}, \cite{sun2017beyond}, \cite{wang2018learning}, \cite{zhao2017spindle} indeed introduce a possible solution for partial biometric identification by dividing an image into multiple patches and then fusing patch-to-patch matching. However, these methods may fail because of requiring the presence of certain person components and pre-alignment. To address the problem of alignment, human parsing, mask \cite{kalayeh2018human}, \cite{qi2018maskreid}, \cite{song2018mask} and skeleton \cite{liu2018pose}, \cite{qian2017pose}, \cite{su2017pose}, \cite{suh2018part} in person re-identification, landmarks in face recognition as external cues are widely used to align persons/faces. But, over-reliance on external cues would result in the biometric system to be unstable in real-world scenarios. Thus, it can be seen that image alignment is a crucial problem for partial biometric identification.

In this paper, we propose a new general robust framework as shown in Fig.~\ref{fig2} for biometric matching that addresses all problems mentioned above and gets rid of fixed inputs on multiple partial biometric identification tasks. In the proposed framework, Fully Convolutional Network (FCN) is utilized to generate spatial feature maps of a certain size. And then a feature post-processing unit consist of global averaging pooling and pyramid pooling is utilized to produce multi-scale spatial feature to avoid the influence of scale variation. Motivated by the remarkable successes achieved by dictionary learning in face recognition \cite{liao2013partial, wright2009robust, zhang2011sparse}, the Spatial Feature Reconstruction (SFR) makes that each spatial feature in the multi-scale spatial maps of the probe image can be sparsely reconstructed on the basis of multi-scale spatial maps of gallery images. In this manner, the model is independent of the size of images and naturally avoids the time-consuming alignment step. Besides, we introduced an objective function namely batch hard triplet which encourages the reconstruction error of the spatial feature maps extracted from the same identity to be minimized while that of different identities being maximized. Generally, the major contributions of our work are summarized as follows:

\begin{itemize}
  \item We propose a robust biometric matching method based on Spatial Feature Reconstruction (SFR) for biometric identification on partial captured objects, which is alignment-free and flexible to arbitrary-sized/ scale images. Hence, the proposed SFR can work well for partial biometric identification.

  \item Spatial feature reconstruction combined with pyramid pooling and global feature matching makes the SFR more robust to scale various, so as to enhance the performance.

  \item We embed the dictionary learning into batch hard triplet learning in a unified framework, and train an end-to-end deep model through minimizing the reconstruction error for coupled images from the same identity and maximizing that of different identities.

  \item Experimental results demonstrate that the proposed approach achieves impressive results on Market1501 \cite{zheng2015scalable}, CUHK03 \cite{zheng2017pedestrian},  DukeMTMC-reID \cite{zheng2017unlabeled}, Partial-REID \cite{zheng2015partial}, and Partial-iLIDs \cite{zheng2011person}, and CASIA-NIR-Distance databases \cite{he2016multiscale}.
\end{itemize}

The paper is built upon our preliminary work reported in \cite{he2018deep} with following improvements: pyramid pooling layer is added to improve the robustness of scale various, $\ell{}_2$ regularization takes the place of $\ell{}_1$ regularization in spatial reconstruction equation for solving the reconstruction coefficient fast, SFR Embedded batch hard triplet learning is utilized to improve the discriminative of spatial feature instead of pairwise learning, spatial feature reconstruction and global feature matching are fused to improve the model performance and we extend the SFR method for more person re-id datasets such as CUHK03 \cite{zheng2015scalable} and DukeMTMC-reID \cite{zheng2017unlabeled} and partial face datasets CAISA-NIR-Distance \cite{he2016multiscale}, Partial LFW, which shows the strong expansibility of SFR.

The remainder of this paper is organized as follows: In Sec.~2, we review the related work about the existing person re-id and partial person re-id algorithms. Sec.~3 introduces the technical details of spatial feature reconstruction and batch hard triplet SFR learning. Sec.~4 shows the experimental results and analyzes the performance in accuracy. Sec.~5 discuss the advantages and disadvantages of the proposed approach. Finally, we conclude our work in Sec.~6.

\section{Literature Review}
As our approach is expected to settle multiple biometric identification problems yet current existing approaches are specialized to one of person re-identification, partial person re-identification or partial face re-identification, we would love to review some of them and give comparisons in Sec.~4 to show our approach holds state-of-the-art on these problems without any specializing and pre-alignment.
\subsection{Person Re-identification}
\noindent\textbf{Part-based models} \cite{chen2016similarity}, \cite{cheng2016person}, \cite{li2017learning}, \cite{wang2018learning}, \cite{zhao2017spindle}, \cite{sun2017beyond} are widely applied to person re-identification since they could achieve significant performance. Zhao \emph{et al.} \cite{zhao2017spindle} proposed a novel Spindle Net based on human body region guided multi-stage feature decomposition and tree-structured competitive feature fusion. Li \emph{et al.} \cite{li2017learning} design a Multi-Scale Context-Aware Network (MSCAN) to learn powerful features over full body and body parts, which can well capture the local context knowledge by stacking multi-scale convolutions in each layer. Moreover, instead of using predefined rigid parts, they proposed to learn and localize deformable pedestrian parts using Spatial Transformer Networks (STN) with novel spatial constraints, which can release some difficulties, e.g. pose variations and background clutters, in part-based representation. Besides, Sun \emph{et al.} \cite{sun2017beyond} proposed a network named Part-based Convolutional Baseline (PCB) that outputs a convolutional descriptor consisting of several part-level features. PCB is able to lay emphasis on the content consistency within each part. However, these methods require the presence of certain person components and pre-alignment.

\noindent\textbf{Mask-guided models} \cite{kalayeh2018human}, \cite{qi2018maskreid}, \cite{song2018mask} provide a solution for person re-identification. Mask as external cue helps to remove the background clutters in pixel-level and contain body shape information. Song \emph{et al.} \cite{song2018mask} introduced the binary segmentation masks to construct synthetic RGB-Mask pairs as inputs, then they design a mask-guided contrastive attention model (MGCAM) to learn features separately from the body and background regions. Kalayeh \emph{et al.} \cite{kalayeh2018human} proposed a person re-identification model that integrated human semantic parsing in person re-identification. Similar to \cite{song2018mask}, Qi \emph{et al.} \cite{qi2018maskreid} combined source images with person masks as the inputs to remove the appearance variations (illumination, pose, occlusion, etc.). Although mask-guided approaches can achieve satisfying performance, they extremely rely on accurate pedestrian segmentation model, otherwise, it would result in poor performance.

\noindent\textbf{Pose-guided models}  \cite{su2017pose}, \cite{suh2018part}, \cite{liu2018pose}, \cite{qian2017pose} utilize skeleton as a external cue in person re-identification to reduce the part misalignment problem. Each part can be well located using person landmarks. Su \emph{et al.} \cite{su2017pose} proposed a Pose-driven Deep Convolutional (PDC) model to learn improved feature extractors and matching models from end-to-end, PDC can explicitly leverages the human part cues to alleviate the pose variations. Suh \emph{et al.} \cite{suh2018part} proposed a two-stream network that consisted appearance map extraction stream and body part map extraction stream. And then a part-aligned feature map is obtained by a bilinear mapping of the corresponding local appearance and body part descriptors. Except for the person alignment, some works \cite{liu2018pose}, \cite{qian2017pose} proposed pose-transferrable models that combined pose estimation and Generative Adversarial Networks (GAN) to augment training samples. The same as the mask-guided models, pose estimation may fail to work due to the loss of person component and severe occlusions.

\noindent\textbf{Attention-based models} \cite{li2018harmonious}, \cite{si2018dual}, \cite{li2018diversity}, \cite{xu2017jointly}, \cite{zhou2017see} take advantages of attention mechanism to extract more discriminative feature. In fact, attention mechanism is a feature selection approach. Li \emph{et al.} \cite{li2018harmonious} formulated a novel Harmonious Attention CNN (HA-CNN) model for joint learning of soft pixel attention and hard regional attention along with simultaneous optimisation of feature representations, dedicated to optimise person re-id in uncontrolled (misaligned) images. Si \emph{et al.} \cite{si2018dual} proposed a dual attention mechanism, in which both intra-sequence and inter-sequence attention strategies are used for feature refinement and feature-pair alignment, respectively. Besides, attentive spatial-temporal networks \cite{li2018diversity}, \cite{xu2017jointly}, \cite{zhou2017see} are widely used in video-based person re-identification task.

\subsection{Partial Person Re-identification} Partial person re-id has become an emerging problem in video surveillance. To address this problem, many methods \cite{donahue2014decaf, girshick2014rich} warp an arbitrary patch of an image to a fixed-size image, and then extract fixed-length feature vectors for matching. However, such method would result in undesired deformation. Part-based models are considered as a solution to partial person re-id. Patch-to-patch matching strategy is employed to handle occlusions and cases where the target is partially out of the camera's view. Zheng $\emph{et al.}$ \cite{zheng2015partial} proposed a local patch-level matching model called Ambiguity-sensitive Matching Classifier (AMC) based on dictionary learning with explicit patch ambiguity modeling, and introduced a global part-based matching model called Sliding Window Matching (SWM) that can provide complementary spatial layout information. However, the computation cost of AMC+SWM is rather expensive as features are calculated repeatedly without further acceleration.

\subsection{Partial Face Recognition} Many approaches \cite{hu2013robust, liao2013partial, weng2016robust} proposed for solving partial face recognition are keypoint-based. Hu \emph{et al.} \cite{hu2013robust} proposed an approach based on SIFT descriptor \cite{lowe2004distinctive} representation that does not require alignment, and the similarities between a probe patch and each face image in the gallery are computed by the instance-to-class (I2C) distance with the sparse constraint. Liao \emph{et al.} \cite{liao2013partial} proposed an alignment-free approach called multiple key points descriptor SRC (MKD-SRC), where multiple affine invariant key points were extracted for facial features representation and sparse representation based on classification (SRC) \cite{wright2009robust} was used for classification. Weng \emph{et al.} \cite{weng2016robust} proposed a Robust Point Set Matching (RPSM) method based on SIFT descriptor, SURF descriptor \cite{bay2006surf} and LBP \cite{ahonen2006face} histogram for partial face matching. Their approach first aligned the partial faces and then computed the similarity of the partial face and a gallery face image. However, the computational cost of each algorithms is expensive and the required alignment step limits its practical applications. Besides, region-based models \cite{cheheb2017random, gutta2002investigation, neo2010development, ou2017robust, pan2007part, sato1998partial, savvides2006partial} also offered a solution for partial face recognition. They only required face sub-regions as input, such as eye \cite{sato1998partial}, nose \cite{sato1998partial}, half (left or right portion) of the face \cite{gutta2002investigation}, or the periocular region\cite{park2009periocular}. He \emph{et al.} \cite{He_2018_CVPR} proposed a Dynamic Feature Matching (DFM) model and achieves the highest performance (94.96\%)for partial face recognition on CASIA-NIR-Distance database \cite{he2016multiscale}. However, these methods require the presence of certain facial components and pre-alignment.  To this end, we propose an alignment-free partial re-identification algorithm  that achieves better performance with higher computation efficiency.
\section{The Proposed Approach}
We will give a clear explanation of the proposed approach in this section from network definition to loss construction. The code is available on  \url{https://github.com/lingxiao-he/Partial-Person-ReID}.

\subsection{Architecture of Deep Network}
For a quick view, the feature matching process is shown in Fig.~\ref{fig2}. In the proposed network, a Fully Convolution Network (FCN) is adopted to extract spatial features, which are post-processed by a unit consist of two feature extraction branches are implemented: global features are extracted by global average pooling layer (GAP) and multi-scale spatial features are extracted by pyramid pooling layer. Then, multi-scale spatial features are fed to SFR, a dictionary learning based reconstruction mechanism supporting matches on arbitrary sized inputs, in feature matching step. Finally, the matching score equals to the weighted sum of results from global matching and SFR matching.

\subsubsection{FCN Encoder}
Models pre-trained on ImageNet \cite{deng2009imagenet} such as VGG \cite{simonyan2014very}
 and ResNet \cite{He_2016_CVPR} can be viewed as a stack of multi-stage convolution layers and a sequence of fully-connected layers. Here we make use of those convolution layers (FCN) in ResNet as our feature encoder. The parameters of the encoder will be fine-tuned in the training process.

\subsubsection{Feature Representation}
This part introduces the two branches in feature representation step. Basicly, global averaging pooling(GAP) produces one scalar representing the feature of whole picture and pyramid pooling gives a batch of features calculated on different receptive fields, which leads better performance in matching objects in arbitrary size and posture.
\begin{figure}[t]
    \centering
    \includegraphics[width=9cm]{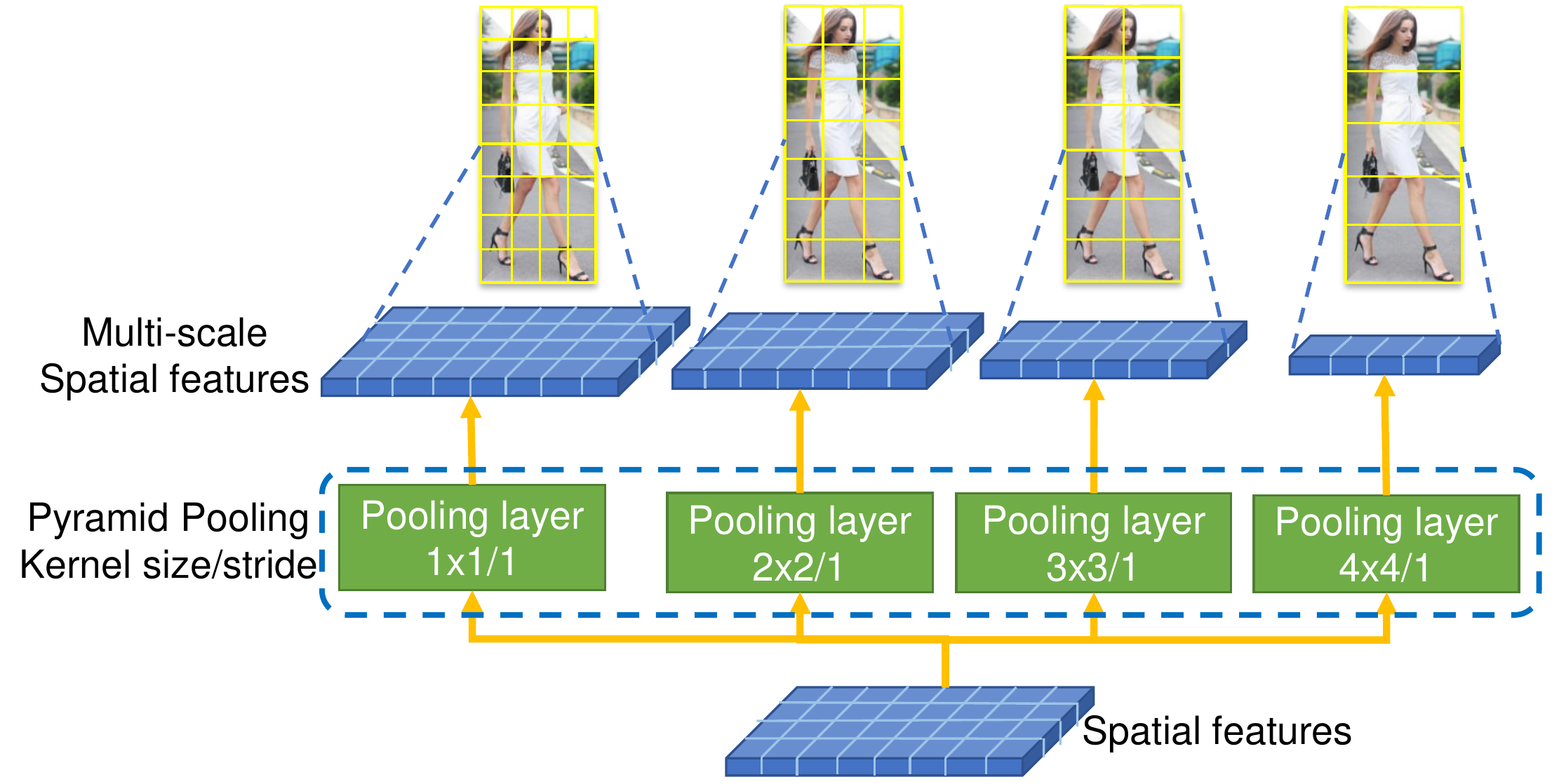}
     \caption{Multi-scale spatial feature representation.}
    \label{fig3}
    \vspace{0em}
\end{figure}

\noindent\textbf{Global Feature}. Global feature is wildly exploited in modern person re-id algorithms. Basicly, Global Averaging Pooling (GAP) realized by a single averaging layer takes the feature maps from FCN as input and outputs one scalar value each image as its global feature. As tested in existing re-id methods that global feature holds relative valid information for matching, we make it in consideration as one of our reference.

\noindent\textbf{Pyramid Feature}. Invariance to varying person scale is a challenging problem for an arbitrary-size person image. It is difficult to align arbitrary-size person image to pre-defined scale. Therefore, the scales between two person images are easily mismatched, resulting in the degraded performance. To this end, we propose pyramid pooling layer to extract multi-scale spatial features to alleviate the influence of scale mismatching.

As shown in Fig.~\ref{fig3}, pyramid pooling (PP) layer consists of multiple average pooling layers of different kernel sizes so that it has different receptive fields. For a $256\times 128$ input person image, we implement 4 pooling layers of different sizes: $1\times 1$, $2\times2$, $3\times 3$, and $4\times 4$ in the pyramid pooling layer. The pyramid pooling layer filters the output spatial features at the stride of 1 to generate multi-scale spatial features.  The output spatial features inferred by pooling layer of small kernel size generate dense spatial features, and each spatial feature represents the local feature of the small local region. The output spatial features inferred by pooling layer of large kernel size generate sparse spatial features, and each spatial feature represents the relatively large source region.  Finally, we concat these output spatial features to obtain multi-scale spatial features. And the multi-scale features are defined as PP$(f_{\theta}(x))$.

\begin{figure}[t]
    \centering
    \includegraphics[width=8.5cm]{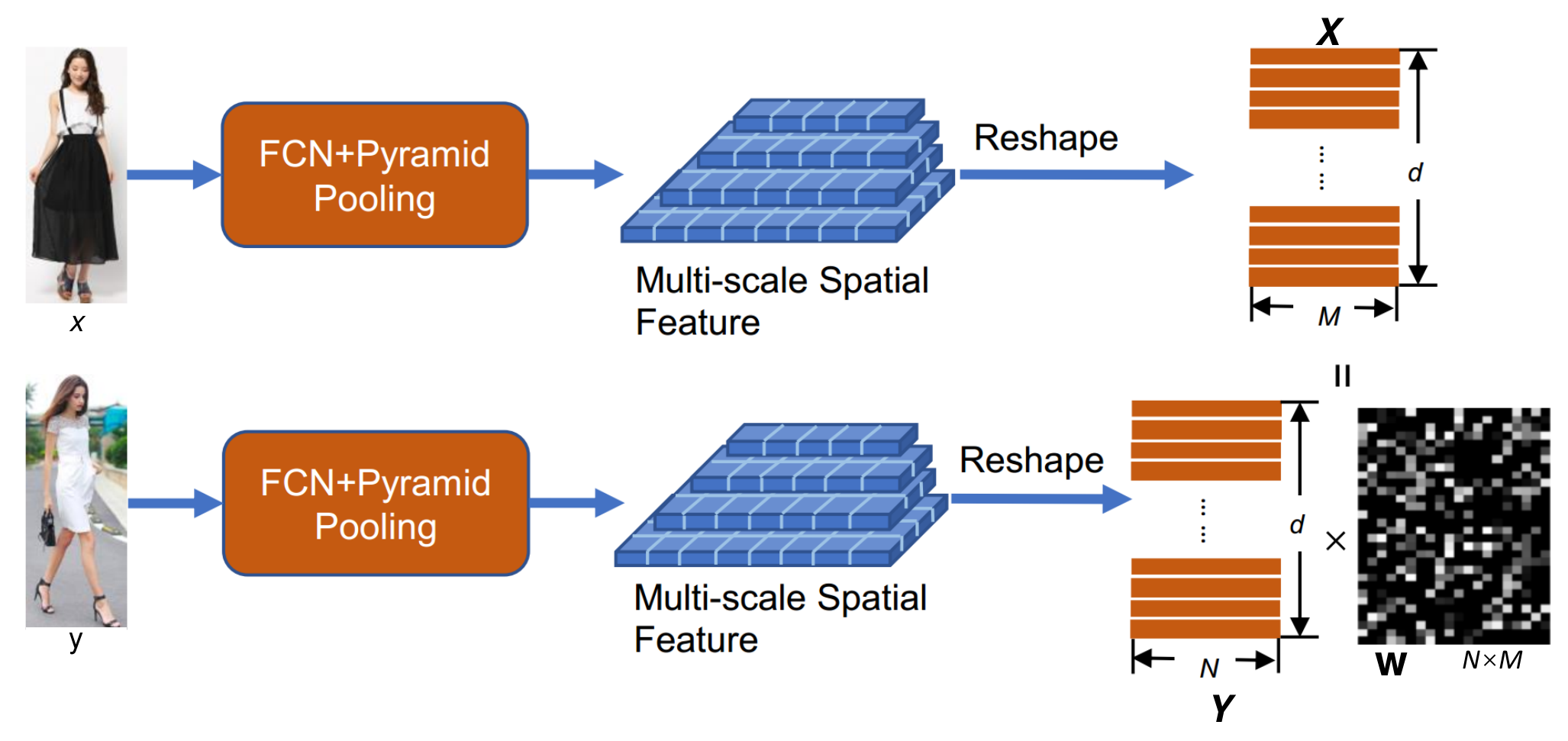}
     \caption{Spatial feature reconstruction.}
    \label{fig4}
\end{figure}
\subsubsection{Spatial Feature Reconstruction}
Spatial feature reconstruction (SFR) between a pair of person images is introduced in this part. As shown in Fig.~\ref{fig4}, for a pair of given person images: $x$ and $y$ with different sizes, correspondingly-size multi-scale spatial features $\mathbf{X} = PP(f_{\theta}(x))=\{\mathbf{x}_1, \ldots, \mathbf{x}_N\}\in \mathbb{R}^{d\times N}$ and $\mathbf{Y}=PP(f_{\theta}(y))=\{\mathbf{y}_1,\ldots,\mathbf{y}_M\} \in \mathbb{R}^{d\times M}$ are then extracted, where $\theta$ denotes the parameters of FCN. Then, $\mathbf{x}_n$ can be represented by linear combination of $\mathbf{Y}$. That is to say, we attempt to search similar spatial features in $\mathbf{Y}$ to reconstruct $\mathbf{x}_n$. Therefore, we wish to solve for the linear representation coefficients $\mathbf{w}_n$ of $\mathbf{x}_n$ with respect to $\mathbf{Y}$, where $\mathbf{w}_n \in \mathbb{R}^{M\times 1}$. We constrain $\mathbf{w}_n$ using $\ell_2$-norm. Then, the linear representation formulation is defined as
\begin{equation}
\begin{array}{l}
 \displaystyle \mathcal{L}(\mathbf{w}_n) = \min_{\mathbf{w}_n}||\mathbf{x}_n-\mathbf{Y}\mathbf{w}_n||_2^{2}+\beta||\mathbf{w}_n||_2,
\end{array}
\label{eq1}
\end{equation}
For $N$ spatial features in $\mathbf{X}$, the Eq. (\ref{eq1}) can be rewritten as
\begin{equation}
\begin{array}{l}
 \displaystyle \mathcal{L}(\mathbf{W}) = \min_{\mathbf{W}}||\mathbf{X}-\mathbf{Y}\mathbf{W}||_2^{2}+\beta||\mathbf{W}||_F,
\end{array}
\label{eq2}
\end{equation}
where $\mathbf{W}=\{\mathbf{w}_1,\ldots,\mathbf{w}_N\}\in \mathbb{R}^{M\times N}$, and $\beta$ controls the smoothness of coding vector $\mathbf{W}$.

We use the least square algorithm to solve $\mathbf{W}$, so $\mathbf{W}=(\mathbf{Y}^{T}\mathbf{Y}+\beta\cdot \mathbf{I})^{-1}\mathbf{Y}^{T}\mathbf{X}$. Let $\mathbf{M} = \mathbf{X}- \mathbf{YW}$, then the spatial feature reconstruction between $\mathbf{X}$ and $\mathbf{Y}$ can be defined as
\begin{equation}
\begin{array}{l}
 \displaystyle D_{s}(\mathbf{X},\mathbf{Y})=tr(\sqrt{\mathbf{M}^{T}\mathbf{M}})/N
\end{array}
\label{eq3}
\end{equation}
where $D_s(:,:)$ is Spatial Feature Reconstruction between a pair of person images.

\begin{algorithm}[t]
\caption{Spatial Feature Reconstruction (SFR).}
\label{alg:Framwork}
\begin{algorithmic}[1] 
\REQUIRE
A probe person image $x$ of an arbitrary-size; a gallery person image $y$.

\ENSURE Similarity score $D_s$. \\ 
\STATE Extract probe multi-scale spatial feature $\mathbf{X}$ and gallery multi-scale spatial feature $\mathbf{Y}$.
\STATE Solve equation (\ref{eq2}) to obtain reconstruction coefficient matrix $\mathbf{W}$.
\STATE Solve equation (\ref{eq3}) to obtain reconstruction score.
\end{algorithmic}
\end{algorithm}
\vspace{1em}
\subsection{Loss Function}
Though pairwise loss with $\ell{}_1$ regularization in our previous work in \cite{he2018deep}, we replace it in this paper by proposed batch hard triple loss with $\ell{}_2$ regularization, which is found performs better than earlier implementation.

\subsubsection{Batch Hard Triplet Loss}
The goal of triplet embedding learning is to learn a function $f_{\theta}(x)$. Here, we want to ensure that an image $x_i^{a}$(anchor) of a specific person is closer to all other images $x_i^{p}$(positive) of the same person than it is to any image $x_i^{n}$(negative) of any other person. Thus, we want $D(x_i^{a}, x_i^{p})+m<D(x_i^{a}, x_i^{n})$, where $D(:,:)$ is Euclidean measure between a pair of person images. So the \emph{Triplet Loss} with $N$ samples is defined as
\begin{equation}
\begin{aligned}
 \displaystyle \mathcal{L}_{tri}(\theta)= \sum_{i}^{N}{[m+D(g_i^{a},g_i^{p})} -D(g_i^{a},g_i^{n})]
\end{aligned}
\end{equation}
where $m$ is a margin that is enforced between positive and negative pairs, and $\mathbf{g}_i^{a}$ = GAP$(f_{\theta}(x_i^{a}))$, $\mathbf{g}_i^{p}$ = GAP($f_{\theta}(x_i^{p})$) and $\mathbf{g}_i^{n}$ = GAP($f_{\theta}(x_i^{n})$).

To effectively select triple samples, batch hard triplet loss modified by triplet loss is adopted: the core idea is to form batches by randomly sampling $P$ subjects, and then randomly sampling $K$ images of each subject, thus resulting in a batch of $PK$ images. Now, for each anchor sample in the batch, we can select the hardest positive and hardest negative samples within the batch when forming the triplets for computing the loss, which is called as \emph{Batch Hard Triplet Loss}:
\begin{equation}
\begin{aligned}
 \displaystyle \mathcal{L}_{BH}(\theta)= \overbrace{\sum_{i=1}^{P}\sum_{a=1}^{K}}^{\text{all anchors}}[m&+\overbrace{\max_{p=1,\ldots,K}D(\mathbf{g}_i^{a},\mathbf{g}_i^{p})}^{\text{hardest positive}}\\ & -\underbrace{\min_{n=1,\ldots,K}D(\mathbf{g}_i^{a},\mathbf{g}_i^{n})}_{\text{hardest negative}}]
\end{aligned}
\end{equation}
which is defined for a mini-batch $\mathcal{B}$ and where a data point $x_i^{j}$ corresponds to the $j$-$th$ image of the $i$-$th$ person in the batch. This results in $PK$ terms contributing to the loss. Additionally,
the selected triplets can be considered moderate triplets, since they are the hardest within a small subset of the data, which is exactly what is best for learning with the triplet loss.
\begin{figure}[t]
    \centering
    \includegraphics[width=8.5cm]{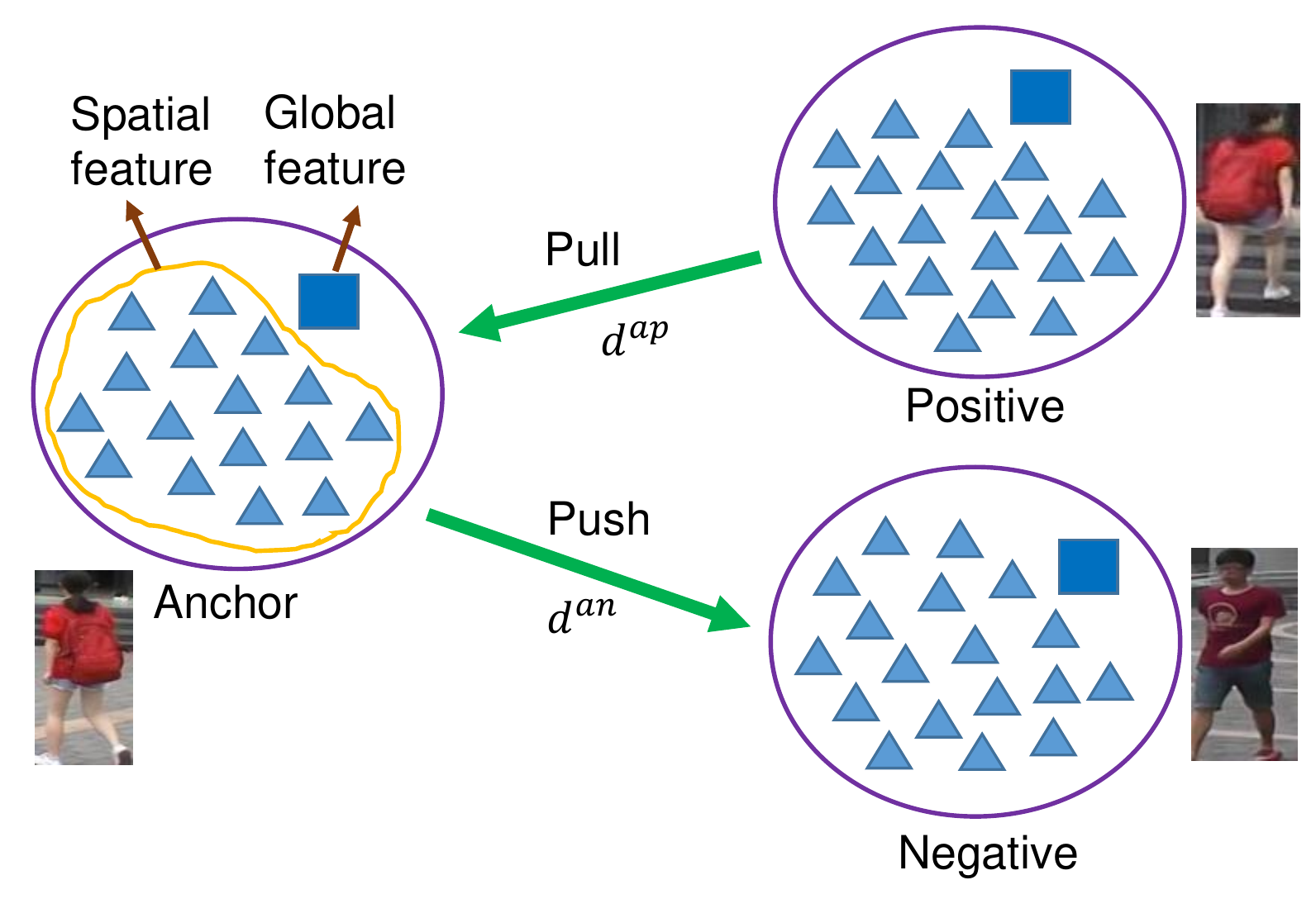}

     \caption{Batch hard triplet SFR learning.}
    \label{fig:fig5}
\end{figure}

\subsubsection{SFR Embedded Batch Hard Triplet}
\emph{Batch Hard Triplet Spatial Feature Reconstruction} is proposed to improve the discriminative of spatial features (see Fig.~\ref{fig:fig5}). It encourages the spatial features of the same identity to be similar while spatial features of the different identities stay away. Batch Hard Triplet Spatial Feature Reconstruction can be defined as
\begin{equation}
\begin{aligned}
 \displaystyle \mathcal{L}(\theta)= \overbrace{\sum_{i=1}^{P}\sum_{a=1}^{K}}^{\text{all anchors}}[m&+\overbrace{\max_{p=1,\ldots,K}(D(\mathbf{g}_i^{a},\mathbf{g}_i^{p})+D_s(\mathbf{X}_i^{a},\mathbf{X}_i^{p}))}^{\text{hardest positive}}\\ & -\underbrace{\min_{n=1,\ldots,K}(D(\mathbf{g}_i^{a},\mathbf{g}_i^{n})+D_s(\mathbf{X}_i^{a},\mathbf{X}_i^{n}))}_{\text{hardest negative}}]
\end{aligned}
\end{equation}
where $D(:,:)$ is Euclidean distance, $D_s(:,:)$ is Spatial Feature Reconstruction distance.

It can be seen that, the similarity distance consists of global feature matching distance (Euclidean distance) and local feature matching distance (spatial feature reconstruction).

\subsubsection{Optimization}
We employ an alternating optimization method to optimize $\theta$.

\noindent\emph{\bf{step 1}: fix $\theta$, obtain $\mathbf{W}_i^{ap}$ and $\mathbf{W}_i^{an}$}. The aim of this step is to solve linear reconstruction coefficient matrix $\mathbf{W}_i^{ap}$ and $\mathbf{W}_i^{an}$
where $\mathbf{W}_i^{ap} =((\mathbf{X}_i^{p})^{T}\mathbf{X}_i^{p}+\beta\cdot I)^{-1}(\mathbf{X}_i^{p})^{T} X_i^{a}$ and $\mathbf{W}_i^{an} =((\mathbf{X}_i^{n})^{T}\mathbf{X}_i^{n}+\beta\cdot \mathbf{I})^{-1}(\mathbf{X}_i^{n})^{T} \mathbf{X}_i^{a}$.

\noindent\emph{\bf{step 2}: fix $\mathbf{W}_i^{ap}$ and $\mathbf{W}_i^{an}$}, optimize $\theta$. We only give the gradients of $D_s(\mathbf{X}_i^{a}, \mathbf{X}_i^{p})$ with respect to $\mathbf{X}_i^{a}$ and $\mathbf{X}_i^{p}$, and the gradients of $D_s(\mathbf{X}_i^{a}, \mathbf{X}_i^{n})$ with respect to $\mathbf{X}_i^{a}$ and $\mathbf{X}_i^{n}$.

\begin{equation}
\begin{aligned}
 \displaystyle \frac{\partial D_s(\mathbf{X}_i^{a}, \mathbf{X}_i^{p})}{\partial{\mathbf{X}_i^{a}}}& = 2 (\mathbf{X}_i^{a}-\mathbf{X}_i^{p}\mathbf{W}_i^{ap})\\
\frac{\partial D_s(\mathbf{X}_i^{a}, \mathbf{X}_i^{p})}{\partial{\mathbf{X}_i^{p}}} &= - 2(\mathbf{X}_i^{a}-\mathbf{X}_i^{p}\mathbf{W}_i^{ap}){\mathbf{W}_i^{ap}}^{T}.\\
\frac{\partial D_s(\mathbf{X}_i^{a}, \mathbf{X}_i^{n})}{\partial{\mathbf{X}_i^{a}}}& = 2(\mathbf{X}_i^{a}-\mathbf{X}_i^{n}\mathbf{W}_i^{an})\\
\frac{\partial D_s(\mathbf{X}_i^{a}, \mathbf{X}_i^{n})}{\partial{\mathbf{X}_i^{n}}} &= - 2(\mathbf{X}_i^{a}-\mathbf{X}_i^{n}\mathbf{W}_i^{an}){\mathbf{W}_i^{an}}^{T}.
\label{eq7}
\end{aligned}
\end{equation}
Then, we use Equation (\ref{eq7}) to compute $\frac{ \partial \mathcal{L}(\mathbf{\theta})}{\partial \mathbf{\theta}}$. Clearly, FCN supervised by SFR is end-to-end trainable and can be optimized by standard Stochastic Gradient Descent (SGD).

\begin{algorithm}[t]
\caption{Feature Learning with SFR Embedded Batch Hard Triplet.}
\label{alg:Framwork1}
\begin{algorithmic}[1] 
\REQUIRE Training data $x_i^{a}, x_i^{p}$ and $x_i^{n}$. The parameter of smoothness strength $\beta$ and learning rate $r$. Pre-trained FCN parameter $\theta$. The total of epoch: T. $t=0$.

\ENSURE FCN parameter $\theta$. \\ 
\STATE \textbf{while} t$<$T \textbf{do}
\STATE Extract multiple spatial feature $\mathbf{X}_i^{a}$, $\mathbf{X}_i^{p}$ and $\mathbf{X}_i^{n}$. And extract global feature $\mathbf{g}_i^{a}$, $\mathbf{g}_i^{p}$ and $\mathbf{g}_i^{n}$.
\STATE $t+1 \leftarrow t$
\STATE Compute the reconstruction error by $\mathcal{L}(\mathbf{\theta})$.
\STATE Update the  sparse reconstruction coefficient matrix $\mathbf{W}_i^{an}$ and $\mathbf{W}_i^{ap}$ using Equation (\ref{eq2}).
\STATE Update the gradients of $\frac{ \partial \mathcal{L}(\mathbf{\theta}^{t})}{\partial \mathbf{\theta}^{t}}$.
\STATE Update the parameters $\mathbf{\theta}$ by $\mathbf{\theta}^{t+1}=\mathbf{\theta}^{t}-r \frac{ \partial \mathcal{L}(\mathbf{\theta}^{t})}{\partial \mathbf{\theta}^{t}}$
\STATE \textbf{end while}
\end{algorithmic}
\end{algorithm}
\subsection{Weighted Feature Matching}

\noindent This subsection will demonstrate the detail of global feature matching, spatial feature reconstruction matching and the weighted fusion of them. Suppose global feature $\mathbf{g}_c$ and spatial feature $\mathbf{Y}_c$ are generated from subject $c$ in the gallery. So the gallery global feature set and spatial feature set are built as respectively:
\begin{equation}
\begin{aligned}
 \displaystyle \text{Global feature set}: \mathbf{G} & = [\mathbf{g}_1, \mathbf{g}_2, \ldots, \mathbf{g}_C]\\
 \text{Spatial feature set}: \mathbf{Y} & = [\mathbf{Y}_1, \mathbf{Y}_2, \ldots, \mathbf{Y}_C]
\end{aligned}
\end{equation}
where $\mathbf{g}_c \in \mathbb{R}^{d}$, $\mathbf{Y}_c \in \mathbb{R}^{k_c \times d}$. $k_c$ is the number of spatial features.
Given an arbitrary-size probe face image, global feature $\mathbf{p}$ and spatial feature $\mathbf{X}$ are generated respectively.
Global feature represents the appearance information of person, we directly use the Euclidean distance: $d_c =||\mathbf{p}-\mathbf{g}_c||_2$ to measure the similarity between two images. Then a distance vector of global feature matching for all the $C$ subjects is denoted as
\begin{equation}
\begin{array}{l}
 \displaystyle \mathbf{d} = \{d_1, d_2, \ldots, d_C\}
\end{array}
\label{eq3}
\end{equation}

Moreover, the spatial feature matching presented above not only capture the spatial layout information of local feature, but it also achieves spatial feature matching without alignment. Therefore, it is robust to pose/view variations and person deformation. Meanwhile, such multi-scale spatial feature representation benefits scale inconsistency. Spatial feature reconstruction can always search similar spatial features from multi-scale spatial feature pool to reconstruct probe spatial feature with minimum error. The spatial feature reconstruction distance is represented as
\begin{equation}
\begin{array}{l}
 \displaystyle r_c = D_s(\mathbf{X}, \mathbf{Y}_c)=tr(\sqrt{\mathbf{M}^{T}\mathbf{M}})/k_c
\end{array}
\label{eq3}
\end{equation}

where $\mathbf{W}_c=(\mathbf{Y}_c^{T}\mathbf{Y}_c+\beta\cdot \mathbf{I})^{-1}\mathbf{Y}_c^{T}\mathbf{X}$, and $\mathbf{M} = \mathbf{X}- \mathbf{Y}_c\mathbf{W}_c$. Then, a distance vector for all the $C$ subjects is denoted as
\begin{equation}
\begin{array}{l}
 \displaystyle \mathbf{r} = [r_1, r_2, \ldots, r_C]
\end{array}
\label{eq3}
\end{equation}

To improve the retrieve accuracy, we combine the two distance vectors. The final distance vector can be written as
\begin{equation}
\begin{array}{l}
 \displaystyle \mathbf{s} = \alpha\cdot \mathbf{d}+(1-\alpha) \cdot\mathbf{r}
\end{array}
\label{eq3}
\end{equation}
where $\alpha$ is a weight for regulating the effect of global feature matching and spatial feature reconstruction. Finally, the identity of the probe image can be determined by $\hat{c}=\arg min_c s_c$, where $s_c$ is the $c^{th}$ entry of $\mathbf{s}$.
\begin{figure}[t]
    \centering
    \includegraphics[width=5.7cm]{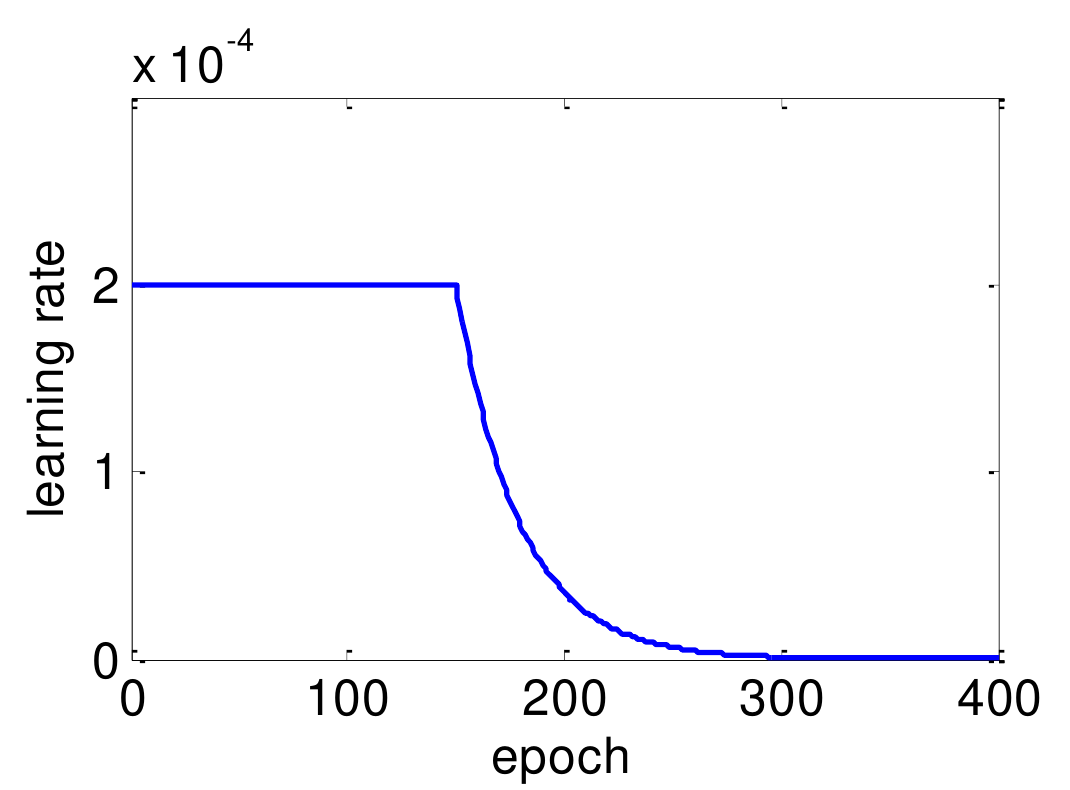}
     \caption{Learning rate curve as a function of the iteration epoch.}
    \label{fig6}
    \vspace{0em}
\end{figure}
\section{Experiments}
To verify the performance as well as the generalization ability of proposed method, this section includes several experiments in the order of person re-identification, partial person re-identification and partial face recognition.
\begin{table*}[t]
  \centering
  \small
  \caption{Performance comparison on Market1501 and CHUK03. R1: rank-1. mAP: mean Accuracy Precision.}

  \label{tab2}
    \begin{tabular}{lllcccccccc}
    \toprule[1.5pt]
    \multicolumn{2}{l}{\multirow{3}{*}{Method}} &
    \multicolumn{4}{c}{Market1501}&\multicolumn{4}{c}{CHUK03} \cr \cline{3-10}
    \multicolumn{2}{c}{~}&\multicolumn{2}{c}{single query}&\multicolumn{2}{c}{multiple query}&\multicolumn{2}{c}{Labeled}&\multicolumn{2}{c}{Detected}\cr\cline{3-10}
     \multicolumn{2}{c}{~}& R1 &mAP &R1 &mAP &R1 &mAP &R1 &mAP  \cr \toprule[1pt]
     \multirow{6}{*}{Part-based}  &Spindle (CVPR17) \cite{zhao2017spindle}&76.50&-&-&-&-&-&-&-\cr
    &MSCAN (CVPR17) \cite{li2017learning}&80.31&57.53&86.79&66.70&-&-&-&- \cr
    &DLPAP (CVPR17) \cite{zhao2017deeply}&81.00&63.40&-&-&-&-&-&-\cr
    &AlignedReID (Arxiv17) \cite{zhang2017alignedreid}&91.80&79.30&-&-&-&-&-&-\cr
    &PCB (Arxiv17) \cite{sun2017beyond}&92.30&77.40&-&-&-&-&61.30&57.50\cr \toprule[1pt]
     \multirow{3}{*}{Mask-guided}  &SPReID (CVPR18) \cite{kalayeh2018human}& 92.54&\bf 81.34&-&-&-&\-&-&-\cr
    &MGCAM (CVPR18) \cite{song2018mask}&83.79 &74.33&-&-&50.14 &50.21&46.71&46.87 \cr
    &MaskReID (Arxiv18) \cite{qi2018maskreid} & 90.02 &75.30 &93.32 &82.29& -&- &- &- \cr \toprule[1pt]
    \multirow{5}{*}{Pose-guided}  &PDC (ICCV17) \cite{su2017pose}&84.14&63.41&-&-&-&-&-&- \cr
    &PABR (Arxiv18) \cite{suh2018part}&90.20&76.00&93.20&82.70&-&-&-&-\cr
    &Pose-transfer (CVPR18) \cite{liu2018pose}&87.65&68.92&-&-&33.80&30.50&30.10&28.20 \cr
    &PN-GAN (Arxiv17) \cite{qian2017pose}&89.43&72.58&-&-&-&-&-&-\cr
    &PSE (CVPR18) \cite{sarfraz2017pose}&87.70&69.00&-&-&-&-&27.30&30.20\cr \toprule[1pt]
    \multirow{3}{*}{Attention-based}  &DuATM (CVPR18) \cite{si2018dual}&91.42&76.62&-&-&-&-&-&- \cr
    &HA-CNN (CVPR18) \cite{li2018harmonious}&91.20&75.70&93.80&82.80&44.40&41.00&41.70&38.60\cr
    &AACN (CVPR18) \cite{xu2018attention}&85.90&66.87&89.78&75.10&-&-&-&- \cr \toprule[1pt]
   \multicolumn{2}{l}{Baseline (ResNet-50+tri)}&88.18&73.85&92.25&80.96&62.14&58.47&60.43&54.24 \cr
   \multicolumn{2}{l}{DSR (CVPR18) \cite{he2018deep}}&91.26&75.62&93.45&82.44&-&-&61.78&56.87 \cr
   \multicolumn{2}{l}{SFR (ours)}&\bf 93.04&  81.02&\bf 94.84& \bf 85.47&\bf 67.29 & \bf 61.47&\bf 63.86&  \bf 58.97\cr\toprule[1.5pt]
    \end{tabular}
\end{table*}
\subsection{Implementation Details and Evaluation Protocol}
Our implementation is based on the publicly available code of PyTorch. All  models in this paper are trained and tested on Linux with GTX TITAN X GPU. In the training term, all training samples are all re-scaled to $256\times 128$, thus $8\times 4$ spatial features are generated by FCN. No data augmentation method is used for training samples. Besides, we set margin $m=0.3$ and $\beta=0.001$ because it can achieve the best performance. With regard to the batch hard triplet SFR function, one batch consists of 32 subjects, and each subject has 4 different images. Therefore, each batch returns 128 groups of hard triples. The model is trained with 400 epochs and the learning rate is shown in Fig.~\ref{fig6}.

For performance evaluation, we employ the standard metrics as in most person ReID literatures, namely the cumulative matching cure (CMC) and the mean Average Precision (mAP). To evaluate our method, we re-implement
the evaluation code provided by \cite{zheng2015scalable} in Python.
\begin{figure}[t]
    \centering
    \includegraphics[width=8.5cm]{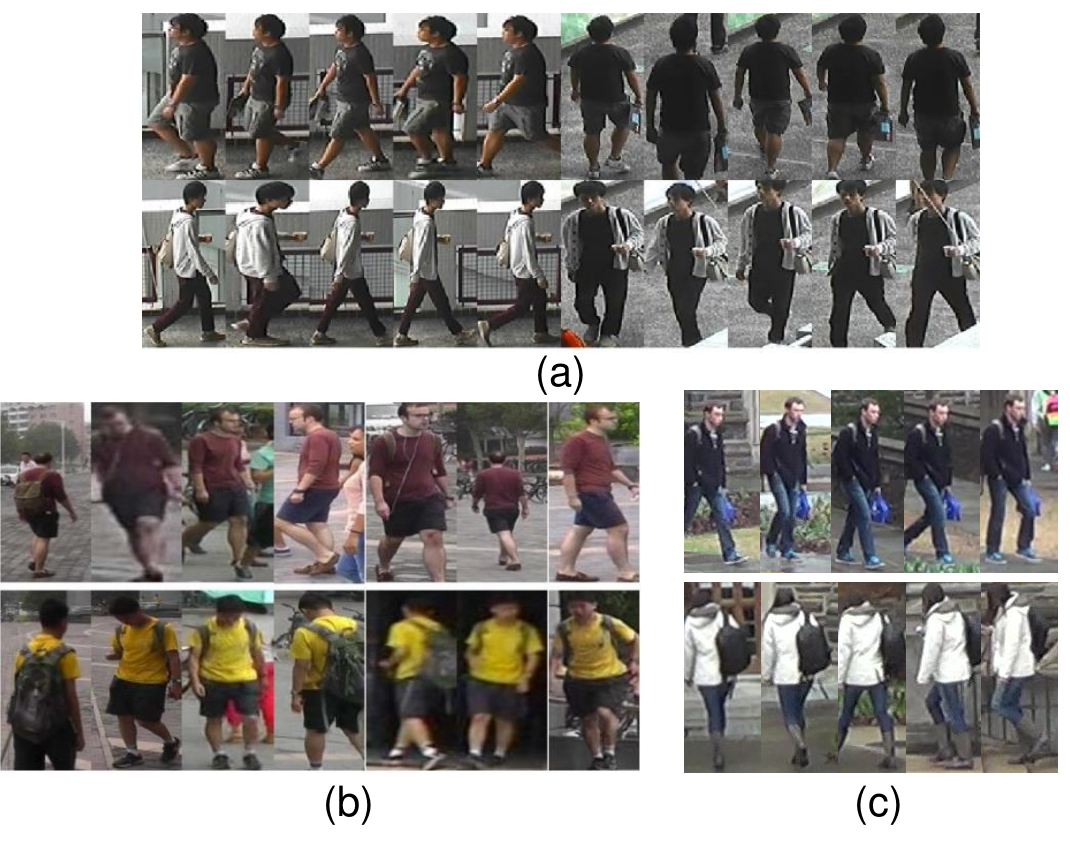}
     \caption{Examples of person images (a) CUHK03 (b) Market1501 (c) Duke.}
    \label{fig7}
    \vspace{0em}
\end{figure}

\subsection{Person Re-identification}
\subsubsection{Datasets}
Three person re-identification datasets: Market1501 \cite{zheng2015scalable}, CHUK03 \cite{zheng2017pedestrian} and DukeMTMC-reID \cite{zheng2017unlabeled} are used for evaluate the proposed SFR.

\noindent\textbf{Market1501} has 12,936 training and 19.732 testing images with 1,501 identities in total from 6 cameras. Deformable Part Model (DPM) is used as the person detector. We follow the standard training and evaluation protocols in \cite{zheng2015scalable} where 751 identities are used for training and the remaining 750 identities for testing.

\noindent\textbf{CHUK03} consists of 13,164 images of 1,467 subjects captured by two cameras from CHUK campus. Both manually labelled and DFM detected person bounding boxes are provided. We adopt the new training/testing protocol \cite{zheng2017pedestrian} proposed in since it defines a more realistic and challenging ReID task. In particular, 767 identities are used for training and the remaining 700 identities are used for testing.

\noindent\textbf{DukeMTMC-reID} is the subset of Duke Dataset \cite{ristani2016performance}, which consists of 16,522 training images from 702 identities, 2,228 query images and 1,7,661 gallery images from the other identities. It provide manually labelled person bounding boxes. Here, we follow the setup in \cite{zheng2017unlabeled}.

The examples of the three datasets are shown in Fig.~\ref{fig7}. And we set $\alpha$\,=\,0.7 in all person re-identification experiments.

\subsubsection{Results}
\noindent\textbf{Results on Market1501}. Comparisons between SFR and 17 state-of-the-art approaches of four categories (part-based model, mask-guided model, pose-guided model and attention-based model) published after 2017 on Market-1501 \cite{zheng2015scalable} are shown in Table~\ref{tab2}. We conduct the single and multiple query experiments, respectively \cite{zheng2015scalable}. The results suggest that the proposed SFR achieves the competitive performance on all evaluation criteria under single and multiple query settings.

It is noted that: (1) The gaps between our results and baseline model (ResNet-50+Triplet) are significant: SFR increases from 88.18\% to 93.04\% under single query setting, and from 92.25\% to 94.84\% under multiple query setting, which fully suggests that spatial feature with alignment-free reconstruction is more effective than only using global feature matching. (2) Benefit from batch hard triplet spatial reconstruction (BHTSR) and pyramid pooling, SFR outperforms our pervious work DSR \cite{he2018deep} by 1.78\%, 1.39\% at the Rank 1 accuracy under single query setting, respectively. BHTSR can learn more discriminative local feature and the pyramid pooling avoids the influence of scale variations of the detected person. (3) Our SFR achieves the best performance at the Rank 1 accuracy. Contributed by exact human semantic parsing, SPReID \cite{kalayeh2018human} achieves the competitive accuracy. However, SPReID relies on excellent human semantic parsing model in a extreme extension and would fail to address arbitrary-size person patch. (4) Although mask and pose estimation provide external cues to improve the performance of person re-identification compared to other methods without using external cues, the overusing of external cues easily result in unstable of these methods due to partial occlusions and the missing of person component. (5) Performance differences among these existing approaches mainly come from input size (e.g., 224 $\times$ 224, 256$\times$ 128 and 384 $\times$ 192), baseline model (e.g., AlexNet, VGGNet, ResNet, and Inception) and algorithms themselves.
\begin{table}[t]
  \small
  \centering
  \caption{Performance comparison on DukeMTMC-reID (R1 means rank score = 1 and mAP: mean Average Precision).}
  \label{tab3}
    \begin{tabular}{llcc}
    \toprule[1.5pt]
   \multicolumn{2}{c}{Method} &R1&mAP  \cr
    \toprule[1pt]
      &Spindle (CVPR17) \cite{zhao2017spindle}&-&-\cr
    Part&MSCAN (CVPR17) \cite{li2017learning}&-&- \cr
    -based&DLPAP (CVPR17) \cite{zhao2017deeply}&-&-\cr
    &AlignedReID (Arxiv17) \cite{zhang2017alignedreid}&-&-\cr
     &PCB (Arxiv17) \cite{sun2017beyond}&81.80&66.10\cr\toprule[1pt]
     Mask-  &SPReID (CVPR18) \cite{kalayeh2018human}&84.43&70.97\cr
    guided&MGCAM (CVPR18) \cite{song2018mask}&- &-\cr
    &MaskReID (Arxiv18) \cite{qi2018maskreid} & 78.86 &61.89  \cr \toprule[1pt]
     &PDC (ICCV17) \cite{su2017pose} &-&- \cr
    Pose- &PABR (Arxiv18) \cite{suh2018part}&84.40&49.30\cr
    guided&Pose-transfer (CVPR18) \cite{liu2018pose}&78.52&56.91 \cr
    &PN-GAN (Arxiv17) \cite{qian2017pose} &73.58&53.20\cr
    &PSE (CVPR18) \cite{sarfraz2017pose}&79.80&62.00\cr \bottomrule[1pt]
    {Attention}  &DuATM (CVPR18) \cite{si2018dual}&81.16&62.26 \cr
    -based&HA-CNN (CVPR18) \cite{li2018harmonious} &80.50&63.80\cr
    &AACN (CVPR18) \cite{xu2018attention}&41.37&-  \cr \bottomrule[1pt]
    \multicolumn{2}{l}{Baseline (ResNet-50+tri)}&80.48&64.80 \cr
    \multicolumn{2}{l}{DSR (CVPR18) \cite{he2018deep}}&82.43&68.73 \cr
    \multicolumn{2}{l}{SFR (ours)}& \bf 84.83& \bf 71.24\cr\toprule[1.5pt]
        \end{tabular}
\end{table}

\noindent\textbf{Results on CUHK03}. We only list the results of those methods that use the new training/testing protocol \cite{zheng2017pedestrian}. Table 6 shows results on CUHK03 when detected person bounding boxes and manually labeled bounding boxes are respectively used for both training and testing. The proposed method SFR get 65.86\% and 63.86\% accuracies while using manually labeled bounding boxes and detected bounding boxes by DPM, respectively. From the results shown in Table~\ref{tab2}, we can find that our proposed method SFR outperforms the previous best method PCB \cite{sun2017beyond} implemented by deep learning with multiple parts by 2.56\% at Rank 1 using detected person bounding boxes. It is also noted that: (1) SFR performs much better than mask-guided model: MGCAM \cite{song2018mask}, pose-guided models: Pose-transfer \cite{liu2018pose} and PSE \cite{sarfraz2017pose}, and attention-based model: HA-CNN \cite{li2018harmonious}. Clear gaps are shown between our method SFR and these state-of-the-art methods: The Rank 1 performance of SFR is 16.00\% higher using either labeled or detected person images than others. The results fully suggest that the advantage of SFR is more pronounced.
(2) Training with BHTSR and multi-scale spatial representation with pyramid pooling performs better than DSR trained with single-scale spatial feature and the pairwise loss function. Similar results are also observed using the mAP metric.

\noindent\textbf{Results on DukeMTMC-reID}. Person Re-ID results on DukeMTMC-reID \cite{zheng2017unlabeled} are given in Table~\ref{tab3}. This dataset is challenging because the person bounding box size varies drastically across different camera views, which naturally suits the proposed SFR with multi-scale representation. Except for Spindle Net \cite{zhao2017spindle}, MSCAN \cite{li2017learning}, DLPAP \cite{zhao2017deeply}, AlignedReID \cite{zhang2017alignedreid}, MGCAM \cite{song2018mask} and PDC \cite{su2017pose}, other comparison methods have reported the results on DukeMTMC-reID. The results show that SFR is 0.40\% and 0.27\% higher than the second best ReID model: SPReID \cite{kalayeh2018human} at the Rank 1 and mAP metrics respectively. Besides, SFR beats the previous work DSR by 2.40\% and 2.51\% at the Rank 1 and mAP metrics, respectively, which indicates that multi-scale representation using pyramid pooling can cope with scale variations to some extent.

\begin{figure}[t]
    \centering
       \vspace{0em}
    \includegraphics[width=9cm]{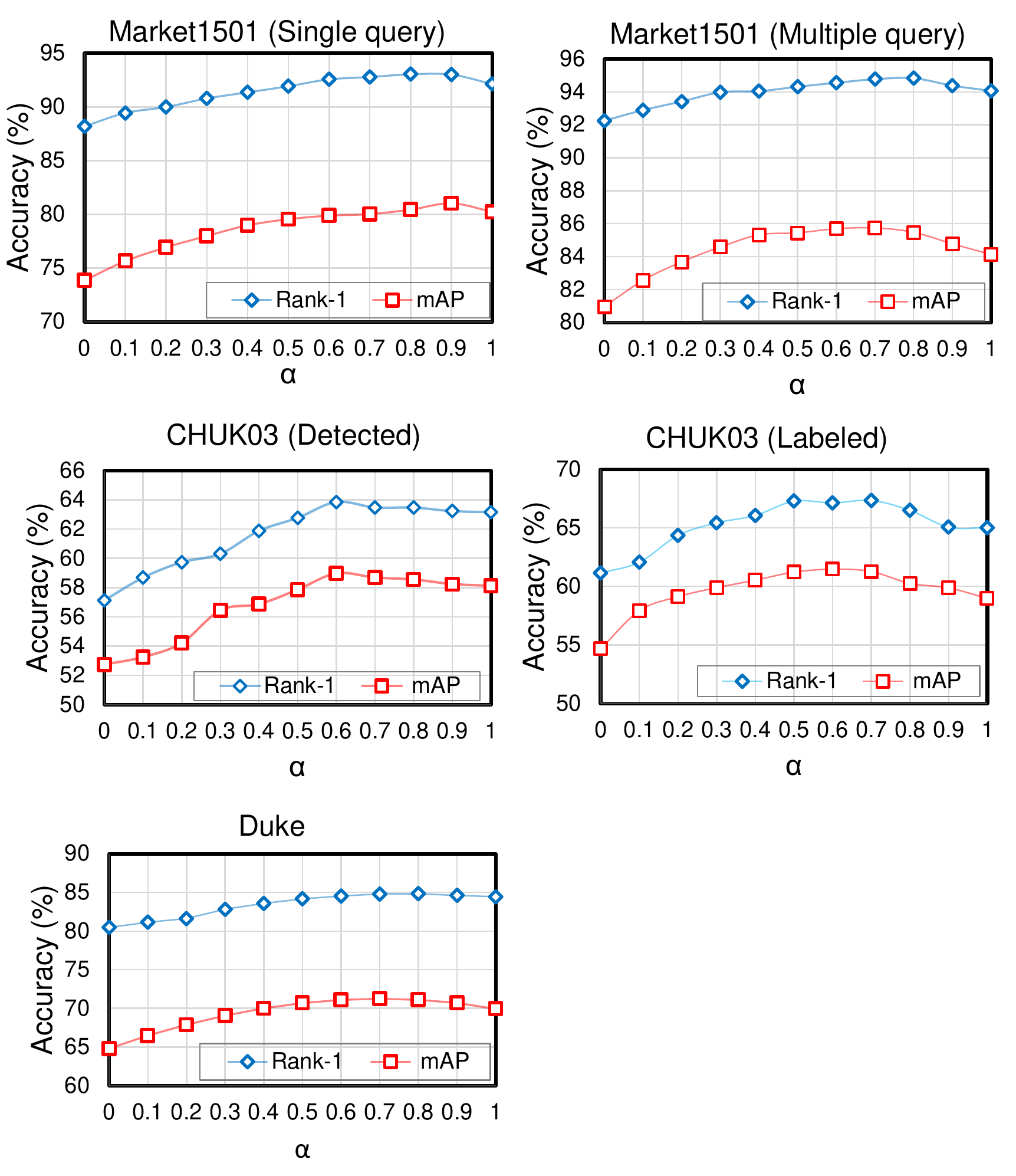}
     \caption{Rank-1 and mAP curves as a function of the weight $\alpha$.}

    \label{fig8}
\end{figure}
\noindent\textbf{Influence of weight $\alpha$}. Similarity measure between two images is achieved by combining global feature matching and spatial feature reconstruction. We set the value of $\alpha$ by from 0 to 1 at the stride of 0.1. Similarity distance only contains global feature matching distance when $\alpha=0$, and similarity distance only contains spatial feature reconstruction when $\alpha=1$. Spatial feature reconstruction performs much better than global feature matching by 3.95\%, 1.82\%, 1.07\%, 0.93\% and 3.95\% on Market1501 under single query and multiple query setting, CHUK03 using labeled and detected person images, and DukeMTMC-reID, respectively. It shows that spatial feature reconstruction is more effective by discovering detail information of the persons. It is worth  to note that fusion of global feature matching and spatial feature reconstruction performs better than single distance measure, which suggests that global feature matching incorporated with spatial feature reconstruction is able to improve the performance of ReID. From the results in Fig.~\ref{fig8}, SFR achieves the best performance when we set $\alpha$\,=\,0.7-0.9, indicating that spatial feature reconstruction is of more importance than global feature matching.

\subsection{Partial Person Re-identification}
\subsubsection{Datasets}
\begin{figure}[t]
    \centering
       \vspace{1em}
    \includegraphics[width=8.8cm]{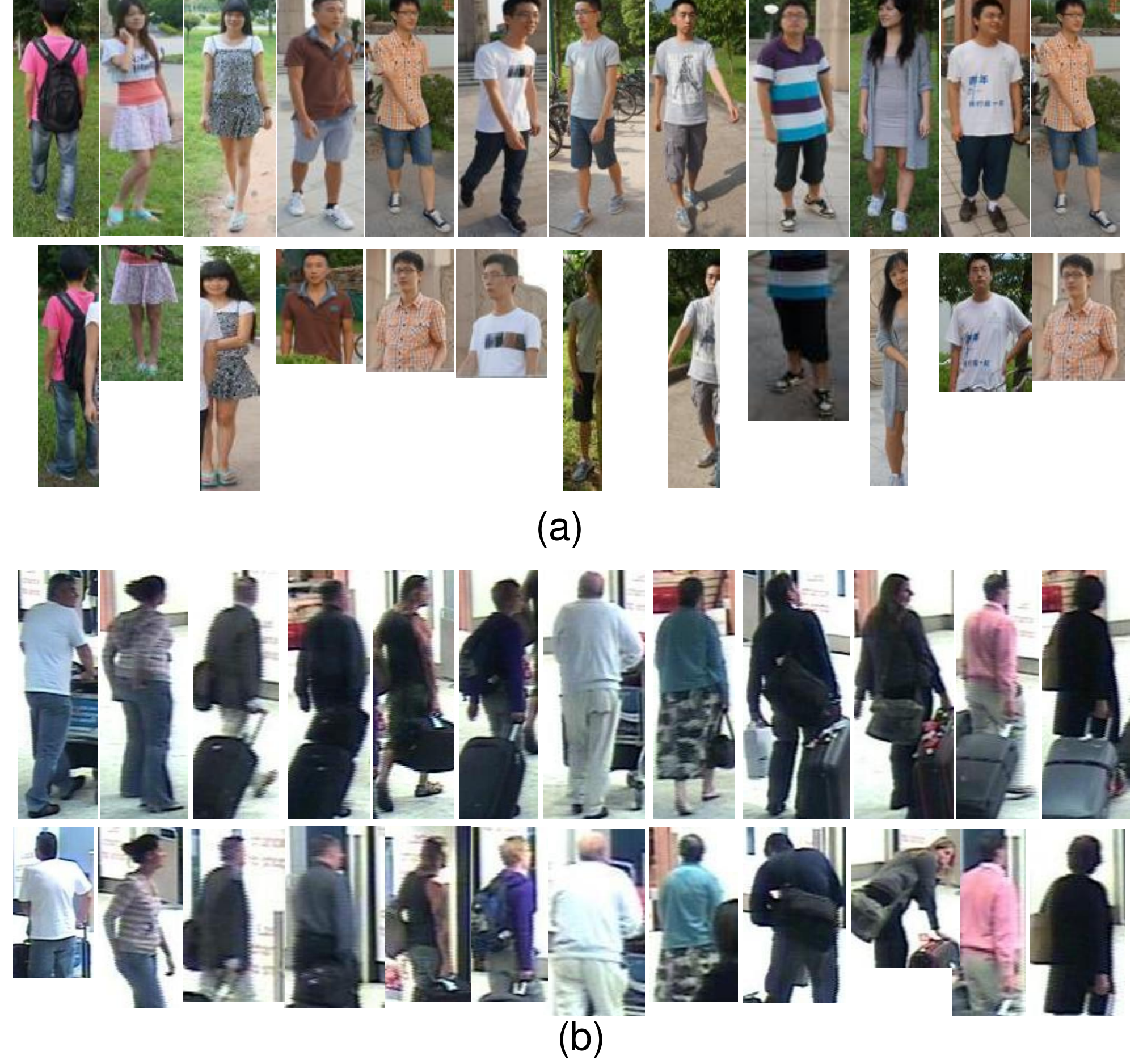}
    \vspace{-1em}
     \caption{Examples of partial persons in Partial REID (a) and P-iLIDS Dataset (b) Datasets.}
    \vspace{0em}
    \label{fig9}

\end{figure}

\noindent\textbf{Partial REID} is a specially designed partial person dataset that includes 600 images from 60 people, with 5 full-body images and 5 partial images per person. These images are collected at a university campus from different viewpoints, backgrounds and different types of severe occlusions. The examples of partial persons in the Partial REID dataset are shown in Fig. \ref{fig9}(a). The region in the red bounding box is the partial person image. The testing protocol can be found in the open code.

\noindent\textbf{Partial-iLIDS} is a simulated partial person dataset based on iLIDS \cite{zheng2011person}. The iLIDS contains a total of 238 images of 119 people captured by multiple non-overlapping cameras. Some images in the dataset contain people occluded by other individuals or luggage. Fig. \ref{fig9}(b) shows some examples of individual images from the iLIDS dataset. For the occluded individuals, the partial observation is generated by cropping the non-occluded region of one image of each person to construct the probe set. The non-occluded images of each person are selected to construct a gallery set.

\begin{table}[t]
\small
  \centering
  \caption{Performance comparison on Partial REID and Partial-iLIDS (Partial images are used as the gallery set and holistic images are used as the probe set).}

    \begin{tabular}{lcccc}
    \toprule[1.5pt]
    &
    \multicolumn{2}{c}{Partial REID}&\multicolumn{2}{c}{Partial-iLIDS}\cr\cline{2-5}
    Rank Score&$1$&$3$&$1$&$3$ \cr
    \toprule[1pt]
    Resizing model &43.87&69.18&26.87&46.22 \cr
    MTSR \cite{liao2013partial} &26.00&37.00&28.57&43.67 \cr
    AMC-SWM \cite{zheng2015partial}&44.67&56.33&52.67&63.33\cr\toprule[1pt]
    Baseline (ResNet-50+tri)& 54.80&80.20 &48.74&68.07\cr
    DSR \cite{he2018deep}& 58.33&82.00 &59.66&78.99\cr
    SFR (ours)& \bf 66.20& \bf 86.67 & \bf 65.55& \bf 81.51\cr\toprule[1.5pt]
    \end{tabular}
    \label{tab4}
\end{table}

\begin{table}[t]
\small
  \centering
  \caption{Performance comparison on Partial REID and Partial-iLIDS (Partial images are used as the probe set and holistic images are used as the gallery set).}
    \begin{tabular}{lcccc}
    \toprule[1.5pt]
    &
    \multicolumn{2}{c}{Partial REID}&\multicolumn{2}{c}{Partial-iLIDS}\cr\cline{2-5}
    Rank Score&$1$&$3$&$1$&$3$ \cr
    \toprule[1pt]
    Resizing model &38.40&56.80&23.67&41.33 \cr
    MTRC \cite{liao2013partial} &23.67&27.33&17.65&26.05 \cr
    AMC+SWM \cite{zheng2015partial}&37.33&46.00&21.01&32.77\cr\toprule[1pt]
    Baseline (ResNet-50+tri)& 43.20&66.33 &42.02&6.87\cr
    DSR \cite{he2018deep}& 50.67&70.33 &58.82&67.23\cr
    SFR (ours)& \bf 56.87& \bf 78.53 &\bf 63.87& \bf 74.79\cr\toprule[1.5pt]
    \end{tabular}
    \label{tab5}
\end{table}
\begin{table}[t]
  \centering
  \small
  \caption{Performance comparison under multi-shot setting on Partial REID.}
  \label{tab6}
    \begin{tabular}{lllcccc}
    \toprule[1.5pt]
        &\multicolumn{2}{c}{N=2}&\multicolumn{2}{c}{N=3}&\cr\cline{2-5}
    Rank Score & $1$ &$3$ &$1$ &$3$   \cr \toprule[1pt]
    Resizing model &46.67&67.33&45.33&67.67\cr
    MTRC \cite{liao2013partial}&33.67&49.67&39.33&57.67 \cr
    AMC+SWM \cite{zheng2015partial}&40.67&58.67&44.67&61.33 \cr \toprule[1pt]

   Baseline (ResNet-50+tri)&59.00&83.33&61.33&84.33 \cr
   DSR (CVPR18) \cite{he2018deep}&69.67&88.33&78.33&88.00 \cr
   SFR (ours)&\bf 73.33& \bf  91.33& \bf 81.33& \bf  92.67\cr\toprule[1.5pt]
    \end{tabular}
    \begin{tabular}{lllcccc}
        &\multicolumn{2}{c}{N=4}&\multicolumn{2}{c}{N=5}&\cr\cline{2-5}
    Rank Score & $1$ &$3$ &$1$ &$3$   \cr \toprule[1pt]
    Resizing model &46.00&68.67&46.33&68.67\cr
    MTRC \cite{liao2013partial}&42.33&61.33&47.67&63.67 \cr
    AMC+SWM \cite{zheng2015partial}&47.67&66.33&50.33&70.67 \cr \toprule[1pt]

   Baseline (ResNet-50+tri)&60.00&84.00&61.33&84.67 \cr
   DSR (CVPR18) \cite{he2018deep}&79.67&91.33&81.00&90.67 \cr
   SFR (ours)& \bf 82.67& \bf  96.00&  \bf 86.33&  \bf 91.33\cr\toprule[1.5pt]
    \end{tabular}
\end{table}

\subsubsection{Results}
The designed Fully Convolutional Network (FCN) is trained with Market1501. We follow the standard training protocols in \cite{zheng2015scalable}, where 751 identities are used for training the FCN model. For comparison, multi-task sparse representation (MTSR) proposed for partial face modeling, ambiguity-sensitive matching and sliding window matching (AMC-SWM) are considered. Besides, Resizing model is also used for comparison, in which person images in the gallery and probe set are all resized to $256\times 128$. And then 2,048-dimension feature vector is extracted by FCN followed by global average pooling (GAP).

\noindent\textbf{Single-Shot Experiments (N=1)}. Single-shot experiment means that only one image per person exists in the probe set. Table \ref{tab4} shows the single-shot experimental results. We find the results on Partial REID and Partial-iLIDS are similar. The proposed method SFR outperforms Resizing model, MTSR, and AMC-SWM. It is noted that:
(1) The gaps between SFR and Resizing model are significant: SFR increases from 43.87\% to 66.20\% and from 26.87\% to 63.87\% at Rank 1 accuracy on Partial REID and Partial-iLIDS, respectively. SFR takes full advantage of FCN that operate in a sliding-window manner and outputs feature maps without deformation. Such results justifies the fact that the person image deformation would significantly affect the recognition performance. For example, resizing the upper part of a person image to a fixed-size would cause the entire image to be stretched and deformed.
(2) AMC-SWM achieves comparable result because local features in AMC combined with global features in SWM makes it robust to occlusions and view/pose various. However, features of non-automatic learning in AMC-SWM make it not as well as SFR performs.
(3) Spatial feature reconstruction combined with global feature matching ($\alpha$=$0.7$ in Partial REID and $\alpha$=$0.6$ in Partial-iLIDS) performs much better than global feature matching (ResNet-50+tri), which fully suggests that the local feature plays a very important role in person re-identification.

\begin{figure}[t]
    \centering
       \vspace{0em}
    \includegraphics[width=9cm]{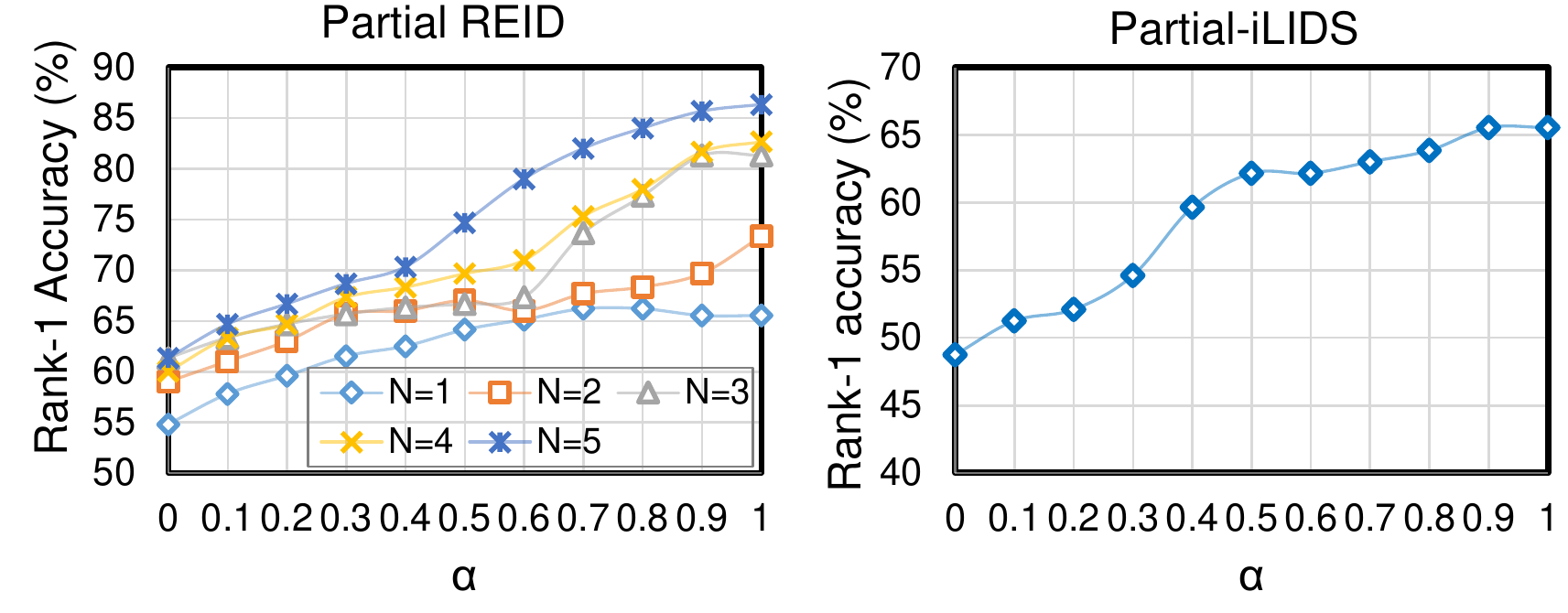}
     \caption{Rank-1 curve as a function of the weight $\alpha$ on Partial REID and Partial-iLIDS.}

    \label{fig11}
\end{figure}

Besides, we conduct another interesting experiment, where we exchange gallery set and probe set. So the gallery set and probe set contain partial person images and holistic person images, respectively. Table~\ref{tab5} shows the experimental result under single-shot settings. Experimental results show that the proposed SFR also performs much better than Resizing model, MTSR, and AMC-SWM and it is also effective when the gallery set only contains partial person images. Furthermore, compared to the results in Table~\ref{tab6}, partial person images exist in the gallery set would influence the performance to some extent.

\noindent\textbf{Multi-shot experiments (N$>$1)}.
Multi-shot means that multiple person images per subject exist in the gallery set. The results are shown in Table~\ref{tab6}. Similar results are obtained in the single-shot experiment, all approaches achieve significant improvement compared to the single-shot experiment. Specifically, the results show that multi-shot setup helps to improve the performance of SFR since it can increase from 66.20\% to 73.33\%, 81.33\%, 82.67\% and 86.33\% on Partial REID dataset at Rank 1 accuracy, respectively.

\noindent\textbf{Influence of weight $\alpha$}.
Similarity measure between two images are achieved by combining global feature matching and spatial feature reconstruction. We set the value of $\alpha$ by from 0 to 1 at the stride of 0.1. Similarity distance only contains global feature matching distance when $\alpha=0$, and similarity distance only contains spatial feature reconstruction when $\alpha=1$. The results are shown in Fig.~\ref{fig11}, we can find that SFR achieves the best rank-1 accuracy under single-shot setting on Partial REID (66.20\%) and Partial-iLIDS ( and 63.87\%) when $\alpha$=0.7 and $\alpha$=0.6, respectively. For multi-shot experiments, we find that SFR performs much better than global feature matching, which can improve more than 10.00\% at the Rank 1 accuracy.
It shows that spatial feature reconstruction is more effective by discovering detail information of the persons.

\begin{figure}[t]
    \centering
       \vspace{1em}
    \includegraphics[width=7cm]{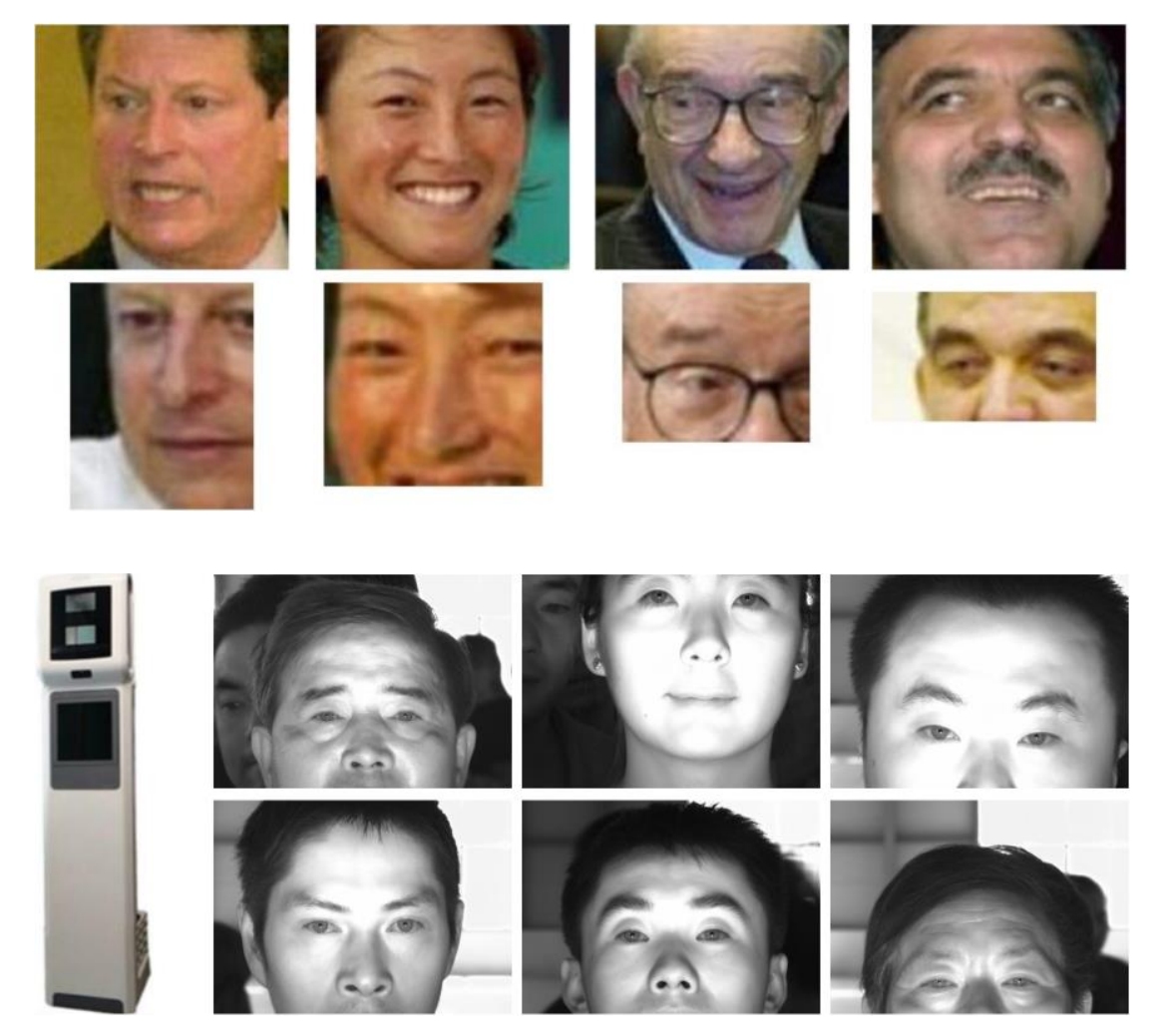}
    \vspace{-1em}
     \caption{Examples of partial face images in Partial LFW (first row) and CASIA-NIR-Distance (second row).}
    \vspace{0em}
    \label{fig12}
\end{figure}
\subsection{Partial Face Re-identification}
\subsubsection{Dataset}
\textbf{CASIA-NIR-Distance} \cite{he2016multiscale} database is a newly proposed partial face database, which contains 4,300 face images from 276 subjects. Half of them contains the entire facial region of the subject. Partial face images are captured by cameras under near-infrared illumination with subject presenting the different arbitrary region of the face. Besides, the variations of presented partial face images in CASIA-NIR-Distance database include imaging at different distances, view angles, scales, and illumination. Fig.~\ref{fig12}(second row) shows some examples of partial faces in the CASIA-NIR-Distance database and the acquisition device.

\noindent\textbf{Partial LFW}, another simulated partial face
database based on LFW database \cite{LFWTech}, is used for evaluation. LFW database contains 13,233 images from 7,749 individuals. Face images in LFW have large variations in pose, illumination, and expression, and may be partially occluded by other faces of individuals, sunglasses, etc.

\subsubsection{Result}
VGGFace \cite{parkhi2015deep} model is used as base model. The fully-connected layers are discarded to evolve into a Fully Convolutional Network (FCN). Close-set experiments are conducted on the CASIA-NIR-Distance and Partial LFW datasets, containing images of 276 and 1,000 subjects respectively. One image per subject (N=1) is selected to construct the gallery set and one different image per subject is used to construct the probe set. For CASIA-NIR-Distance, some subjects do not have holistic face images captured by the iris recognition system, partial face images may exist in the gallery set, thus the difficulty of accurate recognition is increased.  In this experiment, the setting of parameters is that $\alpha$\,=\,0.8. For Partial LFW dataset, the gallery set contains 1,000 holistic face images from 1,000 individuals. The probe set share same subjects with the gallery set, but for each individual they contain different images. Gallery face images are re-scaled to $224\times224$. To generate partial face images as the probe set, an arbitrary-size region at random position of a random size is cropped from a holistic face image. Fig.~\ref{fig12}(first row) shows some partial face images and holistic faces images in Partial LFW.
  \begin{table}[t]
\centering
\small
\caption{Performance comparison on the CASIA-NIR-Distance (p=276, N=1).}

\begin{tabular}{lcccc}
\toprule[1.5pt]

Rank Score& {$1$} & {$3$} & {$5$} & {$10$} \cr
\toprule[1pt]

{MKDSRC-GTP \cite{liao2013partial}}&  83.81  & 85.25 & 86.69& 89.21\\

{RPSM \cite{weng2016robust}}&  77.70  & 80.22 & 82.37& 86.69\\

{I2C \cite{hu2013robust}}&  71.94  & 75.18 & 78.06& 83.81\\
{MRCNN \cite{he2016multiscale}}&  85.97  & 88.13 & 89.93 & 93.17\\
{DFM \cite{He_2018_CVPR}}&  94.96  & 96.40 & 97.84 & 98.55\\ \toprule[1pt]
{SFR }&  \bf 96.74  & \bf 97.46  &\bf 98.55&\bf 99.64 \\ 
\toprule[1.5pt]
\end{tabular}
\vspace{0em}
\label{tab7}
\end{table}
  \begin{table}[t]
\centering
\small
\caption{Performance comparison on the Partial LFW (p=1000, N=1).}

\begin{tabular}{lcccc}

\toprule[1.5pt]

{Rank Score}& {$1$} & {$3$} & {$5$} & {$10$} \cr
\toprule[1pt]

{MKDSRC-GTP \cite{liao2013partial}}&  1.10 & 3.70 & 5.60& 8.40\\
{I2C \cite{hu2013robust}}&  6.80  & 8.30 & 11.20& 14.60\\
{MRCNN \cite{he2016multiscale}}&  24.70  & 28.60 & 31.24 & 35.47\\
{DFM \cite{He_2018_CVPR}}&  27.30  & 34.40 & 39.20 & 47.58\\ \bottomrule[1pt]
{SFR }& \bf 46.30  & \bf 59.30  &\bf 65.50&\bf 70.90 \\
\bottomrule[1.5pt]
\end{tabular}
\vspace{0em}
\label{tab8}
\end{table}

The proposed SFR is compared against the existing partial face algorithms including MRCNN \cite{he2016multiscale}, MKDSRC-GTP \cite{liao2013partial}, RPSM \cite{weng2016robust}, I2C \cite{hu2013robust}, and DFM \cite{He_2018_CVPR}. MKDSRC-GTP, RPSM and DFM are implemented using the source codes provided by authors. I2C is implemented by ourselves according to the paper \cite{hu2013robust}. Table~\ref{tab7} and Table~\ref{tab8} show the performance of the proposed SFR algorithm on the CASIA-NIR-Distance and Partial LFW datasets, respectively. The rank-1 matching accuracies achieved on the two databases are 96.74\% and 46.30\%, which clearly shows that our algorithm performs much better than those traditional algorithms for partial face recognition. The reasons are analyzed as follows:
(1) Multi-scale spatial feature in our SFR takes full advantages of local and global information, which could represent a partial face more robustly in comparison with keypoint-based algorithms (MKDSRC-GTP, RSPM, and I2C).
(2) RPSM method based on SIFT \cite{lowe2004distinctive} descriptor, SURF descriptor \cite{bay2006surf} and LBP \cite{ahonen2006face} histogram for partial face matching first aligns the partial faces and then computes the similarity of the partial face and a gallery face image. However, the required alignment step limits the practical applications of RPSM and the same story happens in MRCNN either.
(3)Although I2C does not require alignment, the similarity between a probe patch and each face image in a gallery is computed by the instance-to-class (I2C) distance with the sparse constraint. Similar to  \cite{hu2013robust}, \cite{weng2016robust}, MKDSRC-GTP simply uses local features and this leads to poor performance.
From these perspectives, the characteristics of alignment-free property and more distinctive and robust descriptions in SFR contribute to the improvement of partial face recognition and place a huge advantage over the existing partial face recognition approaches.

\section{Discussion}
The experiments on person, partial person and partial face re-identificaton  datasets unveil the extensibility of our approach. On each datasets, the proposed approach, SFR, always outperforms other state-of-the-art approaches  including part-based model, mask-guided model, pose-guided model and attention-based model. This is anticipated as these methods require either alignment or external cues, which extremely leads these approaches to poor stability due to relying on segmentation or pose estimation. On the contrast, that SFR relies on both global feature and spatial feature masks it alignment-free, more robust to scale various and external cues unnecessary.

Also, SFR embedded model is able to achieve remarkable performance  without requiring fixed-size input image, which is demanded in AMC-SWM, MTRC and Resizing model. In the form of dictionary learning, SFR is designed for matching a pair of images of different sizes, which makes the model free to address re-id problems of partial images with arbitrary-sizes.

Nevertheless, the proposed approach also has a drawback. Compared to global feature matching, it costs more computational consumption for SFR while solving reconstruction coefficients. Therefore, we are considering the acceleration of the proposed approach as our future work.

\section{Conclusions}
In this paper, we have proposed a novel approach called Spatial Feature Reconstruction (SFR) to get rid of the fixed size input limitation. The proposed spatial feature reconstruction method provides a feasible scheme to reconstruct the probe spatial feature map linearly from a gallery spatial map.
Besides, pyramid pooling layer combined with global pooling layer reduces the influence of scale various, which avoids the alignment step in many other approaches. Furthermore, we embedded SFR into batch hard triplet loss function to learn more discriminative features for minimizing the reconstruction error for a image pair from the same target and maximizing that of image pair from different targets.

Experimental results on several publicly datasets, including Market1501, CUHK03, and DukeMTMC-reIDID datasets, validate the effectiveness and efficiency of SFR. Additionally, the extensibility of the proposed method is unveiled by achieving state-of-the-art results on two partial person datasets: Partial REID and Partial-iLIDS, and on two partial face recognition datasets: CASIA-NIR-Distance and Partial LFW. Finally, the best performance shown on many datasets suggests that combining global feature matching with SFR in this paper is the better solution as it exploits the complementarity of the two feature matching models.


\ifCLASSOPTIONcaptionsoff
  \newpage
\fi

{
\bibliographystyle{ieee}
\bibliography{egbib}
}




\end{document}